\documentclass[letterpaper]{article}
\usepackage{uai2020}
\usepackage[margin=1in]{geometry}

\usepackage{times}

\usepackage[pdftex]{graphicx}
\usepackage[table,dvipsnames]{xcolor}
\usepackage[round]{natbib}
\usepackage[export]{adjustbox}[2011/08/13]

\usepackage{microtype}
\usepackage{booktabs} 

\usepackage{subcaption}

\usepackage{amsmath}
\usepackage{amssymb}
\usepackage{bm}
\usepackage{multirow}
\usepackage{makecell} 

\usepackage[algo2e,ruled]{algorithm2e}

\usepackage[colorlinks]{hyperref}

\hypersetup{citecolor=[rgb]{0.15,0.0,0.8},linkcolor=[rgb]{0.15,0.0,0.8}}

\usepackage{bbold}

\newcommand{\xigone}{\xi_{1}}
\newcommand{\xigtwo}{\xi_{1}}
\newcommand{\xigw}{\xi_{w}}
\newcommand{\omgone}{\omega_{1}}
\newcommand{\omgtwo}{\omega_{2}}
\newcommand{\omgw}{\omega_{w}}

\newcommand{\muf}{\mu_f}
\newcommand{\sigmaf}{\sigma_f}
\newcommand{\sigmagone}{\sigma_{g_1}}
\newcommand{\sigmagtwo}{\sigma_{g_2}}
\newcommand{\sigmagw}{\sigma_{g_w}}
\newcommand{\mugone}{\mu_{g_1}}
\newcommand{\mugtwo}{\mu_{g_2}}
\newcommand{\mugw}{\mu_{g_w}}

\newcommand{\bY}{\mathbf{Y}}
\newcommand{\bA}{\mathbf{A}}
\newcommand{\bG}{\mathbf{G}}
\newcommand{\bSig}{\bm{\Sigma}}
\newcommand{\bgg}{\mathbf{g}}
\newcommand{\bff}{\mathbf{f}}
\newcommand{\bF}{\mathbf{F}}
\newcommand{\bu}{\mathbf{u}}
\newcommand{\by}{\mathbf{y}}
\newcommand{\bx}{\mathbf{x}}
\newcommand{\bX}{\mathbf{X}}
\newcommand{\bz}{\mathbf{z}}
\newcommand{\bk}{\mathbf{k}}
\newcommand{\bK}{\mathbf{K}}
\newcommand{\bKt}{\mathbf{\widetilde{K}}}

\newcommand{\bZ}{\mathbf{Z}}

\newcommand{\EE}{\mathbb{E}}

\newcommand{\NN}{\mathcal{N}}
\newcommand{\LL}{\mathcal{L}}

\newcommand{\KL}{{\rm KL}}
\newcommand{\bmm}{\bm{m}}
\newcommand{\bS}{\mathbf{S}}

\newcommand{\sigmaobs}{\sigma_{\rm obs}}

\newcommand{\betareg}{\beta_{\rm reg}}

\newcommand{\pdspp}{p_{\rm dspp}}

\title{Deep Sigma Point Processes}

\author{ {\bf Martin Jankowiak\thanks{Correspondence to: mjankowi@broadinstitute.org. This work was completed while MJ and JG were at Uber AI.
}} \\The Broad Institute \\
\And
{\bf Geoff Pleiss}  \\ Cornell University \\
\And
{\bf Jacob R. Gardner}   \\ University of Pennsylvania \\}

\begin{document}

\maketitle

\begin{abstract}
We introduce Deep Sigma Point Processes, a class of parametric models inspired by the compositional
structure of Deep Gaussian Processes (DGPs).  Deep Sigma Point Processes (DSPPs) retain
many of the attractive features of (variational) DGPs, including  mini-batch training and predictive uncertainty
that is controlled by kernel
basis functions. Importantly, since DSPPs admit a simple maximum likelihood inference procedure, the
resulting predictive distributions are not degraded by any posterior approximations.
In an extensive empirical comparison on univariate and multivariate regression tasks we
find that the resulting predictive distributions are significantly better calibrated than
those obtained with other probabilistic methods for scalable regression, including variational
DGPs---often by as much as a nat per datapoint.
\end{abstract}

\section{INTRODUCTION}

As machine learning becomes utilized for increasingly high risk applications such as medical treatment suggestion and autonomous driving, calibrated uncertainty estimation becomes critical.
As a result, substantial effort has been devoted to the development of flexible probabilistic models. Gaussian Processes (GPs) have emerged as an important class of models in cases where predictive uncertainty estimates are essential \citep{rasmussen2003gaussian}.
Unfortunately, the simplest GP models often fail to match the expressive power of their neural network counterparts.

This limited flexibility has motivated the introduction of Deep Gaussian Processes (DGPs), which compose multiple layers of latent functions to build up more flexible
function priors \citep{damianou2013deep}. While this class of models has shown considerable promise, inference
remains a serious challenge, with empirical evidence suggesting---not surprisingly---that posterior approximations (e.g. factorizing across layers) can
degrade the performance of the predictive distribution~\citep{havasi2018inference}.
Indeed, targeting posterior approximations as an optimization objective may fail both to achieve good predictive performance and to
faithfully approximate the exact posterior. Clearly, this motivates investigating alternative approaches.

In this work we take a different approach to constructing flexible probabilistic models.
In particular we formulate a class of \emph{fully parametric} models that retain many of the attractive features of variational deep GPs, while posing a much easier (maximum likelihood) inference problem.
This pragmatic approach is motivated by the recognition that---especially in settings where predictive performance is paramount---it is essential to consider the interplay between modeling and inference.
In particular, as motivated above, a compelling class of models may be of limited predictive utility if approximations in the inference procedure severely degrade the posterior predictive distribution.
Conversely, it can be a significant advantage if a class of models admits a simple inference procedure.

Similarly to variational deep GPs, our regression models utilize predictive distributions that are mixtures of Normal distributions whose mean and variance functions
make use of hierarchical composition and kernel interpolation.
In contrast to deep GPs, however, our parametric perspective allows us to directly target the predictive distribution in the training objective.
As we show empirically in Sec.~\ref{sec:exp} the model we introduce---the \emph{Deep Sigma Point Process} (DSPP)---exhibits excellent predictive performance
and dramatically outperforms a number of strong baselines, including Deep Kernel Learning \citep{calandra2016manifold, wilson2016stochastic} and variational Deep Gaussian Processes \citep{salimbeni2017doubly}.

\section{BACKGROUND}
\label{sec:background}

This section is organized as follows.
In Sec.~\ref{sec:gpr}-\ref{sec:sparse} we review the basics of Gaussian Processes and inducing point methods.
In Sec.~\ref{sec:dgp} we review Deep Gaussian Processes.
In Sec.~\ref{sec:ppgpr} we review PPGPR \citep{jankowiak2019sparse}, as it serves as motivation
for DSPPs in Sec.~\ref{sec:dspp}.
We also use this section to establish our notation.

\subsection{GAUSSIAN PROCESS REGRESSION}
\label{sec:gpr}

In probabilistic modeling Gaussian Processes offer flexible non-parametric function priors that are useful in various regression and classification tasks \citep{rasmussen2003gaussian}. For a given input space $\mathbb{R}^d$
GPs are entirely specified by a kernel $k: \mathbb{R}^d \times \mathbb{R}^d \to \mathbb{R}$
and a mean function $\mu: \mathbb{R}^d \to \mathbb{R}$. Different choices of $\mu$ and $k$ permit the modeler to
encode prior information about the generative process.
In the prototypical case of univariate regression\footnote{Note that here and throughout this work we focus on the regression case.}
the joint density takes the form
\begin{equation}
\label{eqn:unireg}
p(\by, \bff | \bX) = p(\by|\bff, \sigmaobs^2) p(\bff | \bX)
\end{equation}
where $\by$ are the real-valued targets, $\bff$ are the latent function values, $\bX = \{ \bx_i \}_{i=1}^N$ are the $N$ inputs with $\bx_i \in \mathbb{R}^d$,
$p(\bff | \bX)$ is a multivariate Normal distribution with covariance $\bK_{NN} = k(\bX,\bX)$,
and $\sigmaobs^2$ is the variance of the Normal likelihood $p(\by|\cdot)$.
The marginal likelihood takes the form
\begin{equation}\label{eqn:marg}
p(\by|\bX) = \int \! d \bff \; p(\by|\bff, \sigmaobs^2) p(\bff | \bX)
\end{equation}
Eqn.~\ref{eqn:marg} can be computed analytically, but doing so is computationally prohibitive for large datasets,
necessitating approximate methods when $N$ is large.

\subsection{SPARSE GAUSSIAN PROCESSES}
\label{sec:sparse}

Recent years have seen significant progress in scaling Gaussian Process inference to large datasets.
This advance has been enabled by the development of inducing point methods \citep{snelson2006sparse,titsias2009variational, hensman2013gaussian}, which we now review. One begins by introducing inducing
variables $\bu$ that depend on
variational parameters $\{ \bz_m \}_{m=1}^{M}$, with each $\bz_m \in \mathbb{R}^d$ and where $M={\rm dim}(\bu) \ll N$.
One then augments the GP prior with the auxiliary variables $\bu$
\begin{equation}\nonumber
p(\bff|\bX) \rightarrow  p(\bff|\bu,\bX,\bZ) p(\bu|\bZ)
\end{equation}
and then appeals to Jensen's inequality to lower bound the log joint density over the inducing variables and the targets:
\begin{align}
\label{eqn:jensenenergy}
\begin{aligned}
\log p(\by, \bu |\bX, \bZ) &= \log \int \! d\bff \, p(\by|\bff) p(\bff|\bu) p(\bu)  \\
&\ge  \EE_{p(\bff|\bu)} \left[ \log p(\by|\bff)  +\log p(\bu) \right] \\
 &=  \sum_{i=1}^N \log \mathcal{N}(y_i | \bk_i^{T} \bK_{MM}^{-1} \bu, \sigmaobs^2)  \\ &\;\;\;\;\; - \tfrac{1}{2\sigmaobs^2}  {\rm Tr} \;\! \bKt_{NN} + \log p(\bu)
 \end{aligned}
 \raisetag{88pt}
\end{align}
where $\bKt_{NN}$ is given by
\begin{equation}
\bKt_{NN} = \bK_{NN} - \bK_{NM}  \bK_{MM} ^{-1} \bK_{MN}
\end{equation}
and
\begin{equation}
\begin{split}
\bK_{MM}&=k(\bZ,\bZ) \qquad \bk_i = k(\bx_i, \bZ) \\
\bK_{NM} &=  \bK_{MN}^{\rm T} = k(\bX,\bZ)
\end{split}
\end{equation}
Eqn.~\ref{eqn:jensenenergy} can be used to construct a variety of algorithms for scalable GP inference; here we limit
our discussion to SVGP \citep{hensman2013gaussian}.

\subsubsection{VARIATIONAL INFERENCE: SVGP}
\label{sec:svgp}

If we apply standard techniques from variational inference to the lower bound in Eqn.~\ref{eqn:jensenenergy},
we obtain SVGP, a popular algorithm for scalable GP inference.
In more detail, SVGP proceeds by introducing a multivariate Normal variational distribution
$q(\bu) = \NN(\bmm, \bS)$
and computing the ELBO, which is the expectation
of Eqn.~\ref{eqn:jensenenergy} w.r.t.~$q(\bu)$ together with an entropy term term $H[q(\bu)]$:
\begin{align}
\begin{aligned}
\label{eqn:svgp}
&\mathcal{L}_{\rm svgp}  = \EE_{q(\bu)} \left[ \log p(\by, \bu |\bX, \bZ) \right] + H[q(\bu)] \\
&= \sum_{i=1}^N \left\{ \log   \mathcal{N}(y_i | \muf(\bx_i), \sigmaobs^2)
    - \frac{1}{2}\frac{\sigmaf(\bx_i)^2}{\sigmaobs^2} \right\}  \\
& -  \KL(q(\bu) | p(\bu))
\end{aligned}
\raisetag{83pt}
\end{align}
Here KL denotes the Kullback-Leibler divergence,
$\muf(\bx_i)$ is the predictive mean function given by
\begin{equation}
\label{eqn:meanfunc}
\muf(\bx_i) = \bk_i^{T} \bK_{MM}^{-1} \bmm
\end{equation}
and $\sigmaf(\bx_i)^2 \equiv \rm{Var}[f_i | \bx_i] $ denotes latent function variance
\begin{equation}
\label{eqn:fvar}
\sigmaf(\bx_i)^2 =  \bKt_{ii} + \bk_i^{T}  \bK_{MM}^{-1} \bS \bK_{MM}^{-1}  \bk_i
\end{equation}
Note that the complete variational distribution used in SVGP is given by
\begin{equation}
    \label{eqn:qdist}
    q(\bff, \bu) = p(\bff | \bu, \bX) q(\bu)
\end{equation}
with a marginal distribution given by
\begin{equation}
    \label{eqn:qfdist}
    q(\bff) = \int \! d\bu \; q(\bff, \bu) = \NN(\bff | \muf(\bX), \bSig_f(\bX))
\end{equation}
where $\bSig_f(\bX)$ is the $N \times N$ covariance matrix
\begin{equation}
    \label{eqn:fcovar}
    \bSig_f(\bX) =  \bKt_{NN} + \bK_{NM}  \bK_{MM}^{-1} \bS \bK_{MM}^{-1}  \bK_{MN}
\end{equation}
The objective $\mathcal{L}_{\rm svgp}$, which depends on $\bmm, \bS, \bZ, \sigmaobs$ and the various kernel hyperparameters,
can then be maximized with gradient methods. Since the expected log likelihood in Eqn.~\ref{eqn:svgp} factorizes as a sum over
data points $(y_i, \bx_i)$ it is amenable to stochastic gradient methods, and thus SVGP can be applied to very large datasets.

\begin{figure*}[t!]
    \begin{minipage}[c]{0.57\textwidth}
  \includegraphics[width=0.99\textwidth,center]{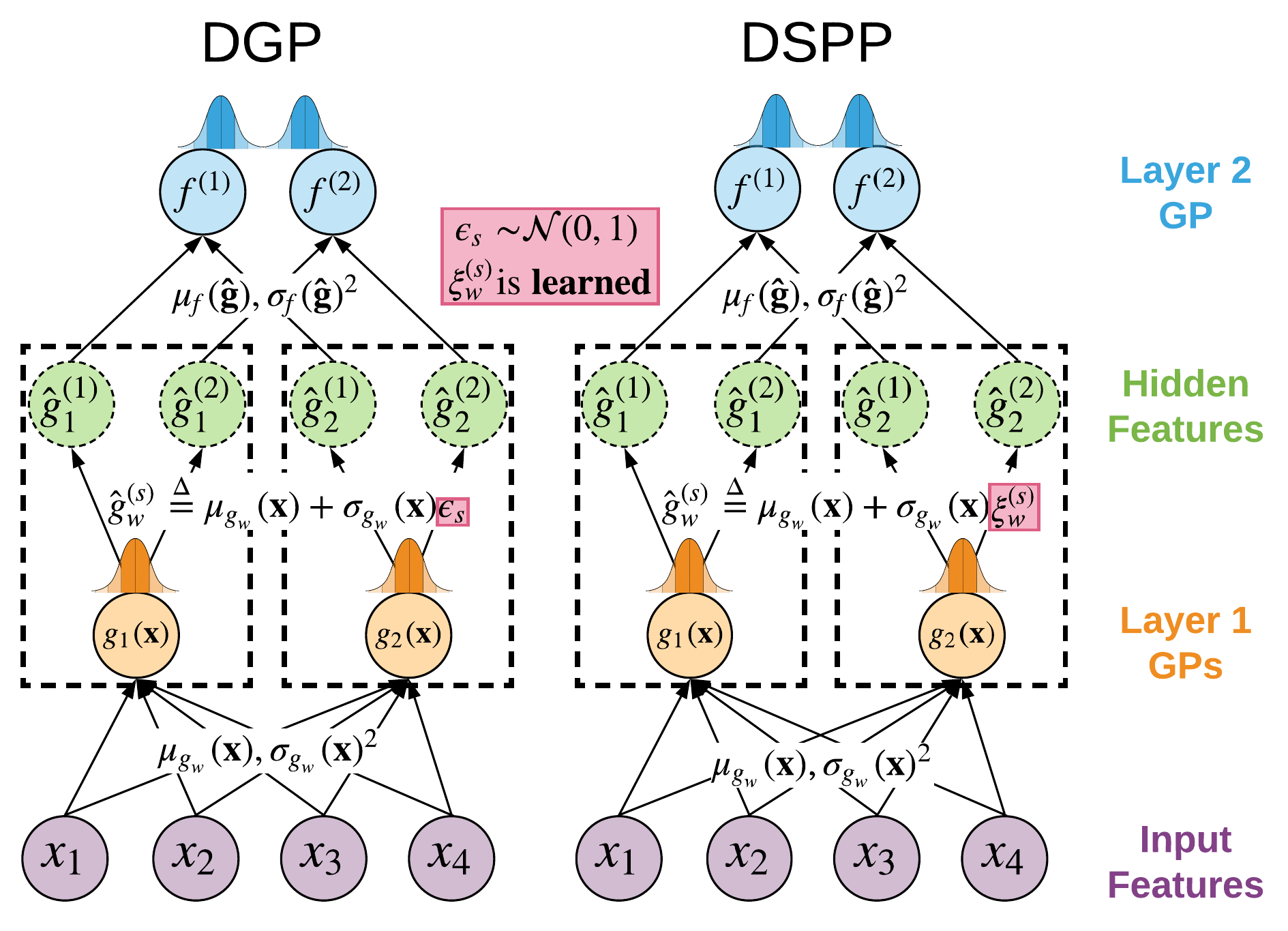}
    \end{minipage}\hfill
    \begin{minipage}[c]{0.42\textwidth}
    \caption{We depict the computional flow of a 2-layer DGP (left; see Sec.~\ref{sec:dgp}) and
             a 2-layer DSPP (right; see Sec.~\ref{sec:dspp}). Input features $\bx$
             are fed through a set of $W$ (here $W=2$) Gaussian Processes
             by computing $\mu_{g_{w}}(\bx)$ and $\sigma_{g_{w}}(\bx)$ (Eqns.~\ref{eqn:meanfunc} and \ref{eqn:fvar}).
             Hidden features for the next GP layer ($f$) are computed from each $\mu_{g_{w}}(\bx)$ and $\sigma_{g_{w}}(\bx)$
             by sampling from $\mathcal{N}(\mu_{g_{w}}(\bx), \sigma_{g_{w}}(\bx))$ in the DGP case,
             or by applying one of the quadrature rules discussed in Sec.~\ref{sec:quadrule} (here QR3),
             with the $\xi_{w}^{(s)}$ treated as learnable parameters.
             The final predictive distribution is a mixture of $S$ Gaussians,
             where each Gaussian depends on one of the $S$ sampled feature sets
             in the DGP case, or on one of the $S$ deterministic quadrature-dependent feature sets in the DSPP case.
             The DGP is trained by gradient descent on $\mathcal{L}_{\rm dsvi}$ (Eqn.~\ref{eqn:dsvi}),
             and the DSPP is trained via $\mathcal{L}_{\rm dspp}$ (Eqn.~\ref{eqn:dsppobj}).
             See Sec.~\ref{sec:dsppdisc} for more discussion on the relationship between DGPs and DSPPs.
    }
     \label{fig:dgpdspp}
    \end{minipage}
\end{figure*}

\subsection{DEEP GAUSSIAN PROCESSES}
\label{sec:dgp}

Deep Gaussian Processes \citep{damianou2013deep} are a natural generalization of GPs in which a sequence
of GP layers form a hierarchical model in which the outputs
of one GP layer become the inputs of the subsequent layer, resulting in a flexible, compositional function prior.
In the simplest case of a 2-layer DGP with a univariate continuous output $y$ and with a GP layer of width $W$
fed into the topmost GP, the joint likelihood for a dataset $(\by, \bX)$ is given by\footnote{We limit our discussion
to this case to simplify notation; generalizations to multiple outputs and multiple layers are straightforward.}
\begin{equation}
\label{eqn:dgp}
    p(\by, \bff, \bG | \bX) = p(\by|\bff, \sigmaobs^2) p(\bff | \bG) p(\bG | \bX)
\end{equation}
Here $\bG$ is a matrix of latent function values of size $N \times W$, i.e.~$g_{iw}$ denotes the latent function value of the $w^{\rm th}$
GP in the first layer evaluated at the $i^{\rm th}$ input $\bx_i$.
For $i=1,...,N$ the vector $\bgg_i \equiv g_{i, :}$ of dimension $W$ then serves as the input to the second GP denoted by $\bff$.
See Fig.~\ref{fig:dgpdspp} for an illustration.
Throughout this work we assume that the prior over $\bG$ factorizes as
$p(\bG | \bX) = \prod_{w=1}^W p(\bgg_w | \bX)$, with each GP governed by its own kernel. Inference in this model
is intractable, necessitating approximate methods. A popular approach is variational inference, which we now briefly review.

\subsubsection{DOUBLY STOCHASTIC VARIATIONAL INFERENCE}
\label{sec:dsvi}

The stochastic variational inference approach described in Sec.~\ref{sec:svgp} can be generalized to the DGP setting
\citep{salimbeni2017doubly}.
Proceeding in analogy to SVGP, we introduce inducing variables and inducing point locations for each GP and
form a factorized variational distribution $Q(\bff, \bu_f, ..., \bgg_W, \bu_{g_W})$ with each factor of the form in
Eqn.~\ref{eqn:qdist}. The variational distribution $Q$ can then be used to form the ELBO
\begin{align}
\begin{aligned}
\label{eqn:dsvi}
    &\mathcal{L}_{\rm dsvi}  = \EE_{Q} \left[ \log p(\by | \bff, \sigmaobs^2) \right] - \sum \; {\rm KL}
\end{aligned}
\raisetag{83pt}
\end{align}
where $\sum \; {\rm KL}$ denotes a sum over the various KL divergences for the
inducing variables $\{\bu_f, ..., \bu_{g_W}\}$.
Crucially, $Q$ is easy to sample from since it suffices to sample from the various
marginals $\{ q(f_i), ..., q(g_{iw}) \}$ and so the expected log likelihood term in Eqn.~\ref{eqn:dsvi} factorizes\footnote{Since e.g.~$\bSig_f(\bX)_{ii} = \sigmaf(\bx_i)^2$ only depends on $\bx_i$.} into a sum over data points $(y_i, \bx_i)$, enabling mini-batch training.
As in SVGP, the topmost layer of latent function values $\bff$
can be integrated out analytically. All remaining latent variables (namely $\bG$) must be sampled, necessitating
the use of the reparameterization trick \citep{price1958useful,salimans2013fixed} and resulting in `doubly' stochastic gradients.
For more details we refer the reader to \citep{salimbeni2017doubly}.

\subsubsection{DGP PREDICTIVE DISTRIBUTIONS}
\label{sec:dgppreddist}

We review the structure of the posterior predictive distribution that results from the variational inference
procedure described in the previous section, as this will serve as motivation for the introduction of Deep Sigma
Point Processes in Sec.~\ref{sec:dspp}.
We first note that in the (single-layer) SVGP case the predictive distribution at input $\bx_*$ is given by
the Normal distribution
\begin{equation}
\label{eqn:svgppreddist}
p(y_* | \bx_*) = \NN( y_* |  \muf(\bx_*), \sigmaf(\bx_*)^2 + \sigmaobs^2)
\end{equation}
where $\muf(\bx_*)$ is the predictive mean function in Eqn.~\ref{eqn:meanfunc} and
$\sigmaf(\bx_*)^2$ is the latent function variance in Eqn.~\ref{eqn:fvar}.
The predictive distribution for the DGP is an immediate generalization of Eqn.~\ref{eqn:svgppreddist}. In
particular in the DGP case the predictive distribution is given by a \emph{continuous mixture} of
Normal distributions of the form in Eqn.~\ref{eqn:svgppreddist}. In more detail, for the 2-layer DGP in Eqn.~\ref{eqn:dgp},
the predictive distribution is given by
\begin{equation}
\label{eqn:dgppreddist}
    \EE_{\prod_{w=1}^W q(g_{*w} | \bx_*)}
    \! \left[ \NN( y_* |  \muf(\bgg_*), \sigmaf(\bgg_*)^2 \! + \! \sigmaobs^2) \right]
\end{equation}
Since this expectation is intractable, in practice it must be approximated with Monte Carlo samples, i.e.~as a finite
mixture of Normal distributions.

\subsection{PPGPR}
\label{sec:ppgpr}

In this section we review a recently introduced approach to scalable GP regression (PPGPR; \citet{jankowiak2019sparse}), as it will
serve as motivation for Deep Sigma Point Processes in Sec.~\ref{sec:dspp}.
In PPGPR one takes the family of predictive distribution in Eqn.~\ref{eqn:svgppreddist}---parameterized by $\bmm$, $\bS$, $\bZ$,
and the kernel hyperparameters---as the model class, and fits the model using gradient-based maximum likelihood
estimation. \citet{jankowiak2019sparse} argue that this class of models achieves good performance because
the training objective directly targets the distribution used to make predictions at test time.
In more detail the PPGPR objective is given by
\begin{equation}
\label{eqn:ppgprobj}
    \begin{split}
        \LL_{\rm ppgpr} = &\sum_{i=1}^N \log p(y_i | \bx_i) - \betareg \KL(q(\bu) | p(\bu)) \\
        = &\sum_{i=1}^N \log   \mathcal{N}(y_i | \muf(\bx_i), \sigmaf(\bx_i)^2 + \sigmaobs^2) \\
        &-\betareg \KL(q(\bu) | p(\bu))
\end{split}
\end{equation}
where $\betareg > 0$ is a hyperparameter and $\KL(q(\bu) | p(\bu))$ serves as a regularizer.
Note that the objective in Eqn.~\ref{eqn:ppgprobj} looks deceptively similar to the SVGP
objective in Eqn.~\ref{eqn:svgp}; the crucial difference is where $\sigmaf(\bx_i)^2$ appears.
This difference is a result of the fact that the PPGPR objective directly targets the predictive distribution,
while the SVGP objective targets minimizing a KL divergence between the variational distribution and
the exact posterior.
In SVGP the posterior predictive is then formed in a second step and does not directly enter the training procedure.
PPGPR will serve as one of our baselines in Sec.~\ref{sec:exp}.

\section{DEEP SIGMA POINT PROCESSES}
\label{sec:dspp}

We now describe the class of models that is the focus of this work. Our approach
is motivated by the good empirical performance exhibited by PPGPR, a class of parametric GP
models explored in \citep{jankowiak2019sparse} and reviewed in Sec.~\ref{sec:ppgpr}.
Crucially, this approach to scalable GP regression relies on an objective function that is formulated
in terms of the predictive distribution. We would like to apply this approach to the DGP setting but, unfortunately, the form
of the predictive distribution for DGPs (see Eqn.~\ref{eqn:dgppreddist}) presents an immediate obstacle:
Eqn.~\ref{eqn:dgppreddist} is a continuous mixture of Normal distributions and cannot be computed in closed form.
In other words, the analog of the PPGPR objective in Eqn.~\ref{eqn:ppgprobj} would involve the logarithm
of the expectation in Eqn.~\ref{eqn:dgppreddist}. While this quantity could be approximated with a Monte Carlo
estimator, because of the outer logarithm the result would be a biased estimator.
Instead of trying to address this obstacle directly and construct an unbiased (gradient) estimator, we instead adopt a simpler solution:
we replace the continuous mixture with a (parametric) \emph{finite} mixture.\footnote{In Sec.~\ref{sec:additional} we report
empirical results comparing to the `direct' (biased) approach in which the predictive distribution is given as a continuous mixture. We find that
this model variant is outperformed by the DSPP, presumably because of the additional model flexibility that results from the learned quadrature rules we adopt in 
Sec.~\ref{sec:quadrule}.}

To define a finite parametric family of mixture distributions, we essentially apply
a sigma point approximation or quadrature-like integration rule to Eqn.~\ref{eqn:dgppreddist}.
To make this concrete, suppose the width of the first GP layer in Eqn.~\ref{eqn:dgppreddist} is $W=2$.
Then for an input $\bx_i$ we can write
\begin{equation}
    \label{eqn:2mix}
\begin{split}
    \pdspp(y_i|\bx_i) = 
    \!\int \! \! d\bgg_{i}
    \NN(y_i| \muf(\bgg_{i}), \sigmaf(\bgg_{i})^2 \!+\! \sigmaobs^2)
    \prod_{w=1}^2 q(g_{iw} | \bx_i) 
\end{split}
\end{equation}
The simplest ansatz for converting Eqn.~\ref{eqn:2mix} into a finite mixture uses Gauss-Hermite quadrature,
i.e.~we approximate $q(g_{i1} | \bx_i)$ with a $S$-component mixture of Dirac delta distributions
controlled by weights $\omgone^{(s)}$ and quadrature points $\xigone^{(s)}$
\begin{equation}
    \label{eqn:gausshermite}
    q(g_{i1} | \bx_i) =  \sum_{s_1=1}^S \omgone^{(s_1)}
    \delta \left(g_{i1} - \left( \mugone(\bx_i) + \xigone^{(s_1)} \sigmagone(\bx_i)  \right) \right)
\end{equation}
with an analogous ansatz for $ q(g_{i2} | \bx_i)$.
For example for $S=3$ we would have
\begin{equation}
\begin{split}
&\xi^{(1)} = -\sqrt{3} \qquad
\xi^{(2)} = 0 \qquad
\xi^{(3)} = -\sqrt{3}  \\
&\omega^{(1)} = \frac{1}{6} \qquad
\omega^{(2)} = \frac{2}{3} \qquad
\omega^{(3)} = \frac{1}{6}
\end{split}
\end{equation}
for both $g_{i1}$ and $g_{i2}$. Making these replacements
in Eqn.~\ref{eqn:2mix} then yields a mixture with $S^2$ components:
\begin{equation}
    \label{eqn:2mixquad}
\begin{split}
    &\pdspp(y_i | \bx_i) = \sum_{s_1}^S \sum_{s_2}^S
    \omgone^{(s_1)} \omgtwo^{(s_2)} \times \\
    &\NN(y_i| \muf(\mugone(\bx_i) + \xigone^{(s_1)} \sigmagone(\bx_i), 
                        \mugtwo(\bx_i) + \xigtwo^{(s_2)} \sigmagtwo(\bx_i) ), \\
            &\;\;\;\;\;\;\;\;\;\sigmaf(\mugone(\bx_i) + \xigone^{(s_1)} \sigmagone(\bx_i),
                        \mugtwo(\bx_i) + \xigtwo^{(s_2)} \sigmagtwo(\bx_i) ))
\end{split}
 \raisetag{60pt}
\end{equation}
Thus for a 2-layer model where the first GP layer has width $W$ this particular quadrature rule
leads to a predictive distribution
that is a mixture of $S^W$ Normal distributions, each of which is parameterized by compositition
of mean and variance functions of the form in Eqn.~\ref{eqn:meanfunc} and Eqn.~\ref{eqn:fvar}.
This exponential growth in the number of mixture components is potentially problematic; we defer a more detailed discussion
of alternative---in particular more compact---quadrature rules to Sec.~\ref{sec:quadrule}.

\subsection{TRAINING OBJECTIVE}
\label{sec:objective}

Now that we have defined the class of parametric regression models we are interested in, we can define
our training objective. As in \citet{jankowiak2019sparse}, we define an objective function that corresponds
to regularized maximum likelihood estimation
\begin{align}
\label{eqn:dsppobj}
    \LL_{\rm dspp} = & \sum_{i=1}^N \log \pdspp(y_i | \bx_i) - \betareg \sum\, \KL
\end{align}
where $\betareg > 0$ is an optional regularization constant and $\sum\, \KL$ is a sum
over KL divergences of inducing variables (one for each GP) just as in the DGP objective, Eqn.~\ref{eqn:dsvi}.
Just as in `Doubly Stochastic Variational Inference' for DGPs (Sec.~\ref{sec:dsvi}), this objective---which depends on
$\sigmaobs$ as well as parameters $\bmm, \bS, \bZ$ and various kernel hyperparameters for each GP---can
be optimized using stochastic gradient methods. In contrast to DGPs, this optimization is only `singly stochastic,'
i.e.~we only subsample data points and not latent function values.

\subsection{QUADRATURE RULES}
\label{sec:quadrule}

The Gauss-Hermite quadrature rule used to motivate DSPPs in Eqn.~\ref{eqn:gausshermite} is intuitive, but is of limited practical use,
since it leads to an exponential blow-up in the number of mixture components used to define the DSPP.
It is therefore essential to consider alternative quadrature rules that are more compact. We emphasize at the outset
that our goal is \emph{not} to accurately estimate the intractable expectation in Eqn.~\ref{eqn:2mix}. Rather, our
goal is to construct a flexible family of parametric distributions that are governed by well-behaved mean and variance functions that benefit
from compositional structure. Consequently, there is no need to restrict ourselves to quadrature rules
derived from gaussian integrals---indeed empirically we find that the quadrature rule in Eqn.~\ref{eqn:gausshermite} is too rigid.
We now describe three more flexible alternatives.
\vspace{-2mm}
\paragraph{QR1}
In the first quadrature rule we choose quadrature points that follow the same factorized structure that is evident
in the double sum in Eqn.~\ref{eqn:2mixquad}, i.e.~we still use a grid of $S^W$ quadrature points for a 2-layer DSPP where
the first GP layer has width $W$. That is we make the substitution
\begin{equation}
\begin{split}
\label{eqn:qr1}
    &\prod_{w=1}^W q(g_{iw} | \bx_i) \rightarrow \\ &\sum_{\bm{s}} \omega^{(\bm{s})}
    \prod_{w=1}^W \delta \left(g_{iw} - \left( \mugw(\bx_i) + \xigw^{(s_w)} \sigmagw(\bx_i)  \right) \right)
\end{split}
 \raisetag{54pt}
\end{equation}
where $\bm{s} = (s_1, ..., s_W) \in \{1, ..., S\}^W$ is a multi-index.
In contrast to the Gauss-Hermite quadrature rule, however, we choose the quadrature
points $\{\xigw^{(s_w)}\}$ to be learnable parameters (for a total of $S \times W$ real parameters).
In addition we replace the Gauss-Hermite weights for each mixture---which
are given as a product over the $W$ GPs in the first layer, i.e. $\prod_w \omgw^{(s_w)}$ for each multi-index
$\bm{s}$---by $S^W$-many learnable parameters that sum to unity $\{ \omega^{(\bm{s})} \}$.
\vspace{-5mm}
\paragraph{QR2}
The second rule is identical to QR1 except the quadrature points are forced to be symmetric,
i.e.~$\xigw^{(s_w)} = - \xigw^{(S+1-s_w)}$ for $s_w=1,...,S$ and $w=1,...,W$.
\vspace{-3mm}
\paragraph{QR3}
In the third quadrature rule we abandon the factorized structure of QR1 and QR2, thus liberating us from the exponential
growth of mixture components. To do this we effectively `line-up' the quadrature points
across the different GPs $g_1, ..., g_W$, making the substitution
\begin{equation}
\begin{split}
\label{eqn:qr3}
    &\prod_{w=1}^W q(g_{iw} | \bx_i) \rightarrow \\ &\sum_{s=1}^S \omega^{(s)} \prod_{w=1}^W
    \delta \left(g_{iw} - \left( \mugw(\bx_i) + \xigw^{(s)} \sigmagw(\bx_i)  \right) \right)
\end{split}
 \raisetag{56pt}
\end{equation}
where there is a set of $S$ learnable quadrature weights $\{ \omega^{(s)} \}_{s=1}^S$ and
$S$ learnable quadrature points $\{ \xigw^{(s)} \}_{s=1}^S$ defined by $S \times W$ real parameters.
Here $S$ is a parameter that we control; in particular it has
no relationship to $W$ and can be as small as $S=1$.
We explore the empirical performance of these quadrature rules in Sec.~\ref{sec:quadruleablation}.

\subsection{DISCUSSION}
\label{sec:dsppdisc}

We use this section to clarify the relationship between variational DGPs and DSPPs.
DSPPs differ from DGPs in two important respects (also see Fig.~\ref{fig:dgpdspp}):
\begin{enumerate}
    \item {\bf Objective function:} The DGP is trained with an ELBO (Eqn.~\ref{eqn:dsvi}) and the DSPP
        is trained via a regularized maximum likelihood objective (Eqn.~\ref{eqn:dsppobj}) that directly
        targets the DSPP predictive distribution.
    \item {\bf Treatment of latent function values:} In the DGP latent function values not at
        the top of the hierarchy (e.g.~$\bG$ in Eqn.~\ref{eqn:dgp}) are \emph{sampled} while in
        the DSPP they are \emph{parameterized} via a learnable quadrature rule as in Eqn.~\ref{eqn:qr3}.
\end{enumerate}
Indeed this latter point is made explicit by our quadrature rules---see Eqn.~\ref{eqn:qr1} and Eqn.~\ref{eqn:qr3}---which should
be compared to the reparameterization trick used during DGP training, which implicitly makes the substitution
\begin{equation}
    \nonumber
    q(g_{iw} | \bx_i) \rightarrow
    \EE_{ \NN(\epsilon | 0, 1)} \Big[ \delta \left(g_{iw} - \left( \mugw(\bx_i) + \epsilon \sigmagw(\bx_i)  \right) \right) \Big]
\end{equation}
and approximates the expectation using Monte Carlo samples $\{\epsilon_s\}$ with $\epsilon_s \sim \NN(\cdot | 0, 1)$.

Note that in other respects variational DGPs and DSPPs are very similar. For example,
apart from the parameters defining the learned quadrature rule in a DSPP, they make use of the
same parameters. Similarly, their predictive distributions have similar forms, with the difference that
the DSPP utilizes a finite mixture of Normal distributions, while the DGP predictive distribution is
a continuous mixture of Normal distributions.

\section{RELATED WORK}
\label{sec:related}

We discuss some work related to DSSPs, noting that we review some of the relevant
literature in Sec.~\ref{sec:background}.
The use of pseudo-inputs and inducing point methods to scale-up Gaussian Process inference has spawned a large literature, especially
in the context of variational inference
\citep{snelson2006sparse,titsias2009variational, hensman2013gaussian, cheng2017variational}.
While variational inference remains the most popular inference algorithm for scalable GPs,
researchers have also explored different variants of Expectation Propagation \citep{hernandez2016scalable} as well as
Stochastic gradient Hamiltonian Monte Carlo \citep{havasi2018inference}.

Deep Gaussian Processes were introduced in \citep{damianou2013deep},
with recent approaches to variational inference for DGPs described by \citet{salimbeni2017doubly}.
\citet{cutajar2017random} introduce a hybrid model formulated with random feature expansions that
combines features of DGPs and neural networks.
This class of models bears some resemblance to DSPPs;
this is especially true for the \texttt{VAR-FIXED} variant, in which spectral frequencies are treated deterministically.
Importantly, since \citet{cutajar2017random} rely on variational inference, their training objective does not directly
target the predictive distribution. As we show empirically in Sec.~\ref{sec:exp}, the mismatch between the training objective
and test time predictive distributions for variational DGPs severely degrades predictive performance; we expect this is also true of the
approach in \citep{cutajar2017random}. 

DSPPs can also be motivated by Direct Loss Minimization, which emerges from a view of approximate inference as regularized loss minimization \citep{sheth2017excess}. This connection is somewhat loose, however, since our fully parametric models dispose of approximate posterior
(or quasi-posterior) distributions entirely.

\section{EXPERIMENTS}
\label{sec:exp}

\begin{table}[t!]
  \centering
    \caption{Average ranking of different quadrature rules (further to the left is better).
  CRPS is the Continuous Ranked Probability Score, a popular calibration metric for regression \citep{gneiting2007strictly}.
    Rankings are aggregated across the smallest 8 UCI datasets and train/test/validation splits.
    See Sec.~\ref{sec:quadruleablation} for details.
    }
  \resizebox{0.85\linewidth}{!}{%
    \begin{tabular}{ccccc}
\toprule
{} &   & DSPP-QR1 & DSPP-QR2 &           DSPP-QR3 \\
\midrule
NLL  &   &   $2.01$ &   $2.44$ &  $\mathbf{ 1.37 }$ \\
RMSE &   &   $1.79$ &   $2.27$ &  $\mathbf{ 1.62 }$ \\
CRPS &   &   $1.73$ &   $2.51$ &  $\mathbf{ 1.51 }$ \\
\bottomrule
\end{tabular}

  }\label{table:uciranks_abliation}
\end{table}
\begin{figure*}[t!]
  \centering
  \includegraphics[width=1.1\textwidth,center]{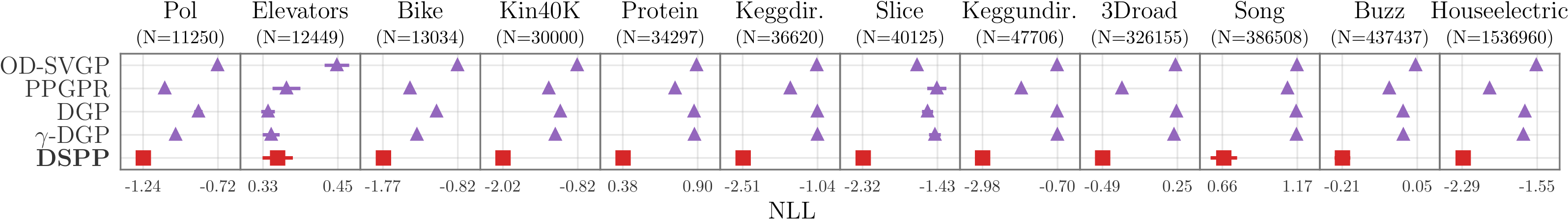}
    \caption{We depict negative log likelihoods (NLLs) for the 12 univariate regression datasets in Sec.~\ref{sec:uci} (further to the left is better).
    Results are averaged over ten random train/test/validation splits.
    Here and throughout uncertainty bars depict standard errors.}
  \label{fig:ucill}
\end{figure*}
\begin{figure*}[t!]
  \centering
  \includegraphics[width=1.1\textwidth,center]{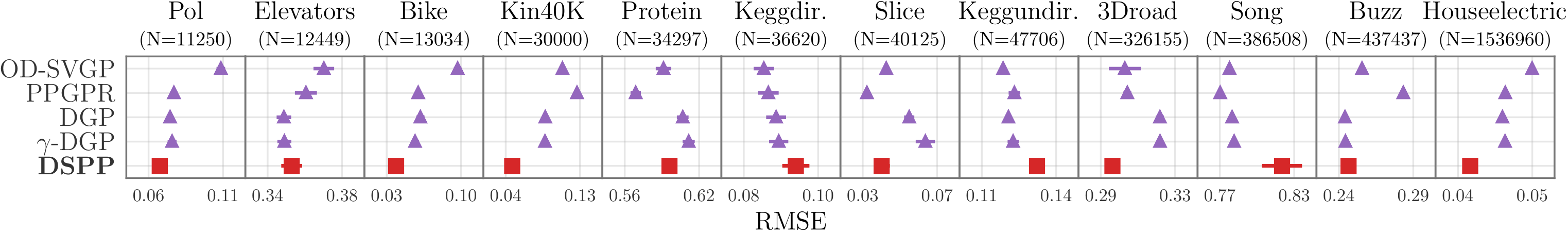}
    \caption{We depict root mean squared errors (RMSEs) for the 12 univariate regression datasets in Sec.~\ref{sec:uci} (further to the left is better).
    Results are averaged over ten random train/test/validation splits.}
  \label{fig:ucirmse}
\end{figure*}
\begin{figure*}[t!]
  \centering
  \includegraphics[width=1.1\textwidth,center]{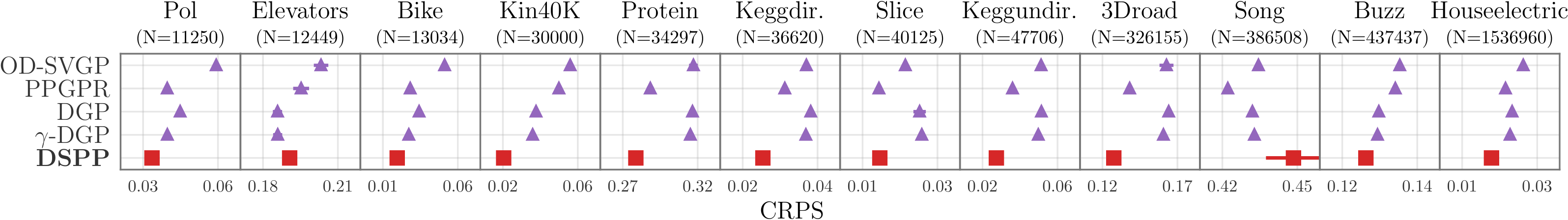}
    \caption{We depict the Continuous Ranked Probabilistic Score (CRPS; \citet{gneiting2007strictly}) for the 12 univariate regression datasets in Sec.~\ref{sec:uci} (further to the left is better). CRPS is a popular calibration metric for regression.
    Results are averaged over ten random train/test/validation splits.}
  \label{fig:ucicrps}
\end{figure*}
\begin{table}[t!]
  \centering
    \caption{Average ranking of methods in Sec.~\ref{sec:uci} (lower is better).
     Rankings are aggregated across all 120 pairs of dataset and train/test/validation split.
    }
  \label{tab:ranks}
  \resizebox{0.95\linewidth}{!}{%
    \begin{tabular}{ccccccc}
\toprule
{} &   & OD-SVGP &   PPGPR &     DGP & $\gamma$-DGP &               DSPP \\
\midrule
NLL  &   &  $4.38$ &  $2.33$ &  $3.55$ &       $3.60$ &  $\mathbf{ 1.15 }$ \\
RMSE &   &  $3.28$ &  $2.85$ &  $2.98$ &       $3.32$ &  $\mathbf{ 2.58 }$ \\
CRPS &   &  $4.42$ &  $2.47$ &  $3.66$ &       $2.93$ &  $\mathbf{ 1.52 }$ \\
\bottomrule
\end{tabular}

    }\label{table:uciranks}
\end{table}

In this section we explore the empirical performance of the Deep Sigma Point Process introduced
in Sec.~\ref{sec:dspp}. Throughout DSPPs use diagonal (i.e.~mean field) covariance matrices $\bS$.
Except for Sec.~\ref{sec:multilayer} where we consider 3-layer models, all DGPs
and DSPPs considered in our experiments have two layers. For details about kernels, mean functions, layer
widths, numbers of inducing points, etc., refer to Sec.~\ref{sec:suppexp} in the appendix.

\subsection{QUADRATURE RULE ABLATION STUDY}
\label{sec:quadruleablation}

We begin by exploring the impact of the three different quadrature rules defined in
Sec.~\ref{sec:quadrule}. To do so we train 2-layer DSPPs on the 8 smallest
univariate regression datasets described in the next section. In particular we compare
QR1 and QR2 with\footnote{Since we consider $W\in\{3,4\}$ this corresponds to
$S^W \in \{27,81\}$ mixture components.} $S=3$ to QR3 with $S=10$. The results are summarized
in Table \ref{table:uciranks_abliation}, with more detailed results to be found in Sec.~\ref{sec:additional} in the appendix.
The upshot is that the three quadrature rules give largely comparable performance, with some preference
for the most flexible quadrature rule, QR3. Since this quadrature rule avoids exponentially many
quadrature points---and the computational cost can be carefully controlled by choosing $S$
appropriately---we use QR3 in the remainder of our experiments. Indeed this choice is essential for training
3-layer DSPPs in Sec.~\ref{sec:multilayer}, which would otherwise be too computationally expensive.

\subsection{UNIVARIATE REGRESSION}
\label{sec:uci}

We reproduce a univariate regression experiment described in \citep{jankowiak2019sparse}.
In particular we consider consider twelve datasets from the UCI repository \citep{Dua:2019},
with the number of data points in the range $10^4 \lessapprox N \lessapprox 10^6$ and the number of
input dimensions in the range $3 \le {\rm Dim}(\bx) \le 380$.

We consider four strong GP baselines---two single-layer models:
({\bf OD-SVGP}) the orthogonal basis decoupling method described in \citet{cheng2017variational} and \citet{salimbeni2018orthogonally};
({\bf PPGPR}) the Parametric Predictive GP regression method described in Sec.~\ref{sec:ppgpr};
as well as two 2-layer models:
({\bf DGP}) a variational DGP as described in Sec.~\ref{sec:dgp}; and
($\bm{\gamma}${\bf -DGP}) the robust DGP described in \citet{knoblauch2019dgp,knoblauch2019generalized}.\footnote{This robust DGP can
be viewed as a variant of the DGP described in Sec.~\ref{sec:dgp} in which the inference procedure is modified by
replacing the expected log likelihood loss with a gamma divergence in which the likelihood is raised
to a power: $\log p(\by|\bff) \rightarrow p(\by|\bff)^{\gamma-1}$.} Results for OD-SVGP and PPGPR
are reproduced from \citep{jankowiak2019sparse}.

Our results are summarized in Figs.~\ref{fig:ucill}-\ref{fig:ucicrps} and Table \ref{table:uciranks}.
We find that in aggregate the DSPP outperforms all the baselines in terms of log likelihood, RMSE, and CRPS (also see
Table \ref{table:multi} in Sec.~\ref{sec:additional} in the appendix).
In particular, averaging across all twelve datasets, the DSPP outperforms the DGP by $\sim 0.75$ nats
and the PPGPR by $\sim 0.47$ nats w.r.t.~log likelihood.
Strikingly, the second strongest baseline is the (single-layer) PPGPR described in Sec.~\ref{sec:ppgpr}. The fact
that the PPGPR is able to outperform a 2-layer DGP highlights the advantages of a training procedure
that directly targets the predictive distribution. Since the DSPP is in effect a much more flexible version
of the PPGPR, it yields even better predictive performance, especially with respect to log likelihood.
Indeed while the DGP achieves good RMSE performance on most datasets, posterior approximations
degrade the calibration of the test time predictive distribution (as measured by log likelihood and CRPS).

\subsection{MULTIVARIATE REGRESSION}
\label{sec:multi}

\begin{figure*}[t!]
  \centering
  \includegraphics[width=0.48\textwidth]{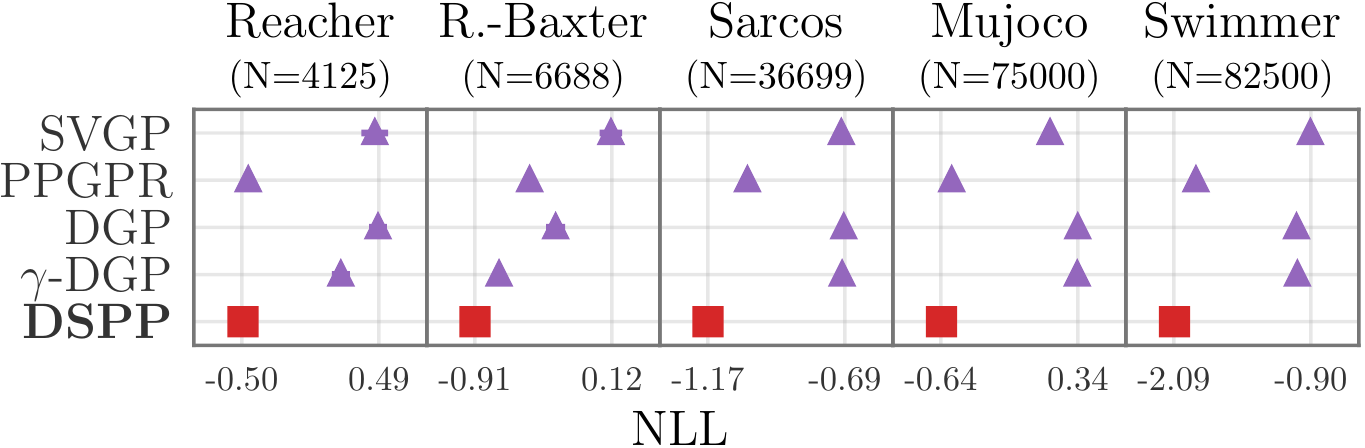}
	\hspace{0.02\textwidth}
  \includegraphics[width=0.48\textwidth]{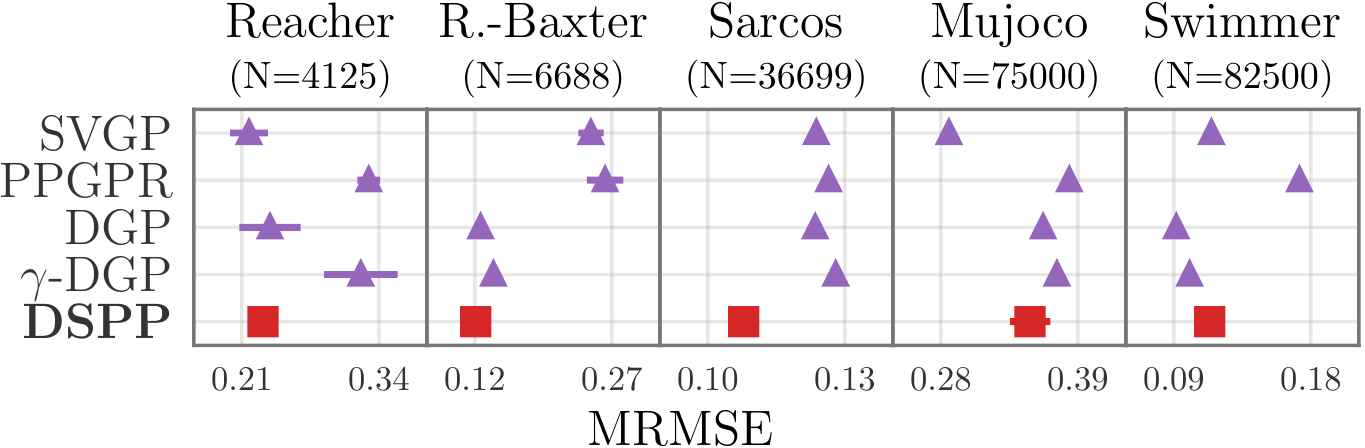}
    \caption{We depict NLLs (left) and mean root mean squared errors (MRMSEs, right)---i.e.~RMSEs averaged across all output dimensions---for the 5 multivariate regression datasets in Sec.~\ref{sec:multi} (lower is better).
    Results are averaged over five random train/test/validation splits.
    Note that NLLs are normalized by the number of output dimensions.
		}
  \label{fig:multill_mrmse}
\end{figure*}
\begin{table}[t!]
  \centering
    \caption{Average ranking of methods in Sec.~\ref{sec:multi} (lower is better).
    Rankings are aggregated across all 25 pairs of dataset and train/test/validation split.
    }
  \label{tab:multi_ranks}
  \resizebox{1.0\linewidth}{!}{%
    \begin{tabular}{ccccccc}
\toprule
{} &   &    SVGP &   PPGPR &                DGP & $\gamma$-DGP &               DSPP \\
\midrule
NLL   &   &  $4.16$ &  $2.20$ &             $4.24$ &       $3.40$ &  $\mathbf{ 1.00 }$ \\
MRMSE &   &  $2.52$ &  $4.72$ &  $\mathbf{ 2.04 }$ &       $3.60$ &             $2.12$ \\
\bottomrule
\end{tabular}

    }\label{table:multiranks}
\end{table}

We conduct a multivariate regression experiment using
five robotics datasets, two of which were collected from real-world robots and three of which were generated
using the MuJoCo physics simulator \citep{todorov2012mujoco}.
In all five datasets the input and output dimensions correspond to various joint positions/velocities/etc.~of the robot,
with the number of data points in the range $10^4 \lessapprox N \lessapprox 10^5$, the number of
input dimensions in the range $10 \le {\rm Dim}(\bx) \le 23$, and
the number of output dimensions in the range $7 \le {\rm Dim}(\by) \le 10$.
These datasets have been used in a number of papers, including 
\citep{vijayakumar2000locally,cheng2017variational}.

All of our models follow the structure of the `linear model of coregionalization' \citep{alvarez2012kernels}, specifically along the lines
of `Semiparametric latent factor models' \citep{seeger2005semiparametric}; for more details see Sec.~\ref{sec:suppmulti} in the supplementary
materials.

We consider four strong GP baselines. In particular we consider two single-layer models:
({\bf SVGP}) as described in Sec.~\ref{sec:svgp} and \citep{hensman2013gaussian};
({\bf PPGPR}) the Parametric Predictive GP regression method described in Sec.~\ref{sec:ppgpr};
as well as two 2-layer models:
({\bf DGP}) a variational DGP as described in Sec.~\ref{sec:dgp}; and
($\bm{\gamma}${\bf -DGP}) the robust DGP described in \citet{knoblauch2019dgp, knoblauch2019generalized}.

Our results are summarized in Fig.~\ref{fig:multill_mrmse} and Table~\ref{table:multiranks}; see Sec.~\ref{sec:additional} in the
appendix for additional results. We find that the DSPP outperforms all the baselines w.r.t.~NLL, and achieves
comparable MRMSE performance to the DGP.
Averaged across all five datasets, the DSPP outperforms the DGP by $\sim 0.82$ nats and PPGPR by $\sim 0.17$ nats
 w.r.t.~log likelihood.
Note that while the PPGPR achieves good performance on log likelihoods,
its MRMSE performance is substantially worse than the DSPP.

\subsection{MULTILAYER MODELS}
\label{sec:multilayer}

\begin{figure*}[t!]
  \centering
  \includegraphics[width=0.47\textwidth]{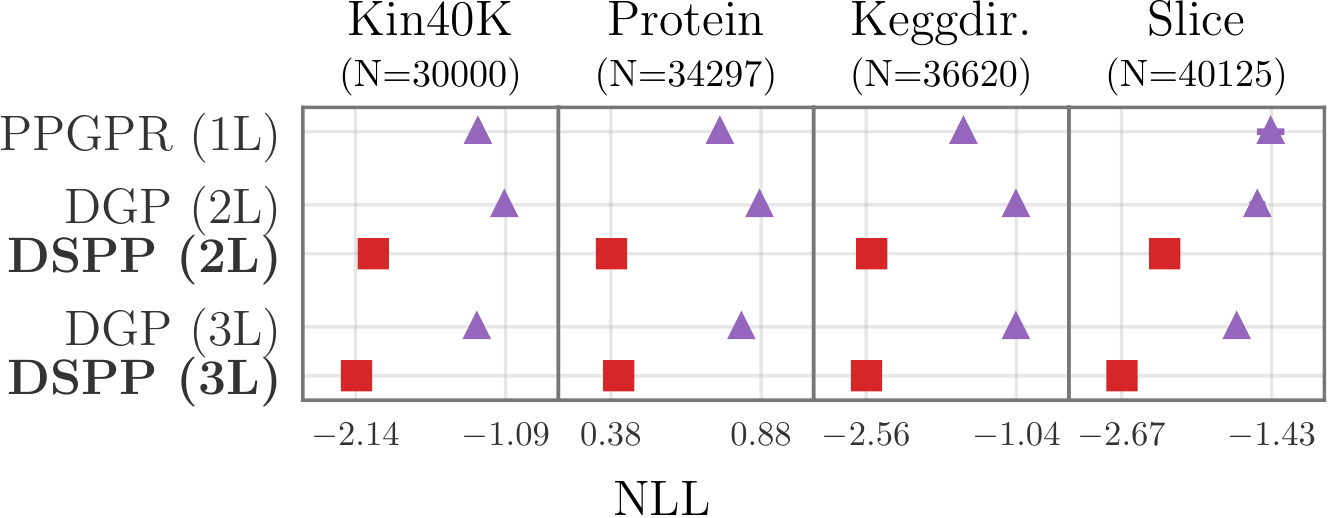}
  \hspace{0.05\linewidth}
  \includegraphics[width=0.47\textwidth]{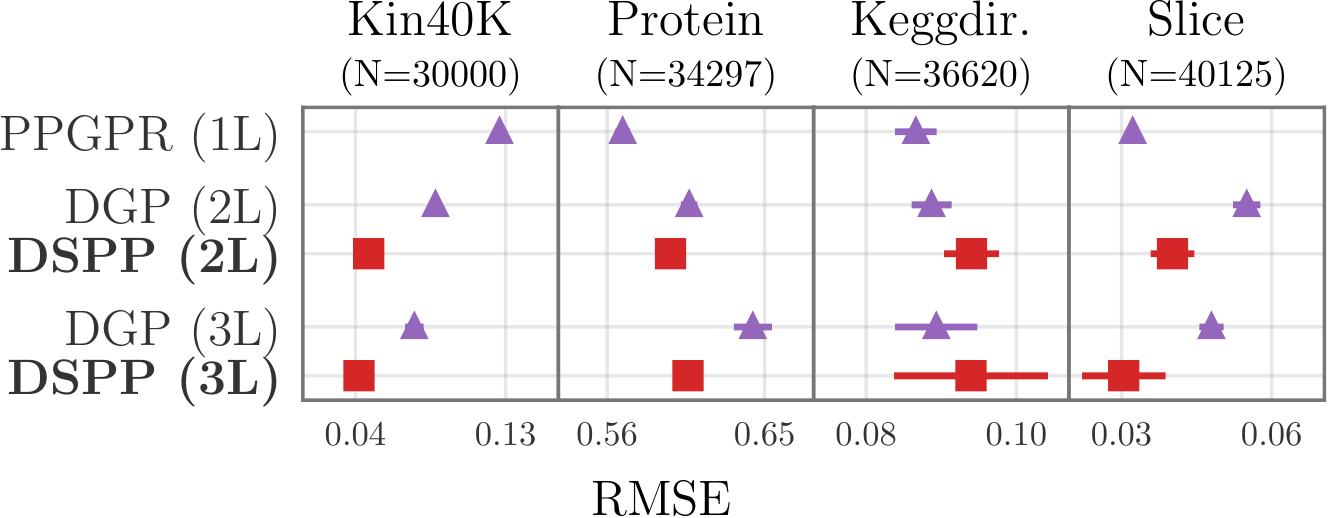}
    \caption{We depict NLLs (left) and RMSEs (right) for the multi-layer experiment
             in Sec.~\ref{sec:multilayer} (lower is better). Results for 1L and 2L (respectively, 3L) models
              are averaged over ten (respectively, five) random train/test/validation splits.}
  \label{fig:threelayerllrmse}
\end{figure*}

In the above experiments we have limited ourselves to 2-layer DSPPs.
Here we investigate the predictive performance of 3-layer models, in particular comparing to 3-layer DGPs.
Our results for four univariate regression datasets are summarized in Fig.~\ref{fig:threelayerllrmse}.
For both DGPs and DSPPs we find that,
depending on the dataset, adding a third layer can improve NLLs, but that these gains are somewhat marginal compared
to the gains from moving from single-layer to two-layer models. DSPPs exhibit the best log likelihoods, while
RMSE results are somewhat more mixed, with the DSPP and PPGPR obtaining the lowest RMSE depending on the dataset.

\subsection{COMPARISON WITH DEEP KERNEL LEARNING}
\label{sec:dkl}

\begin{figure*}[t!]
  \centering
  \includegraphics[width=0.48\linewidth]{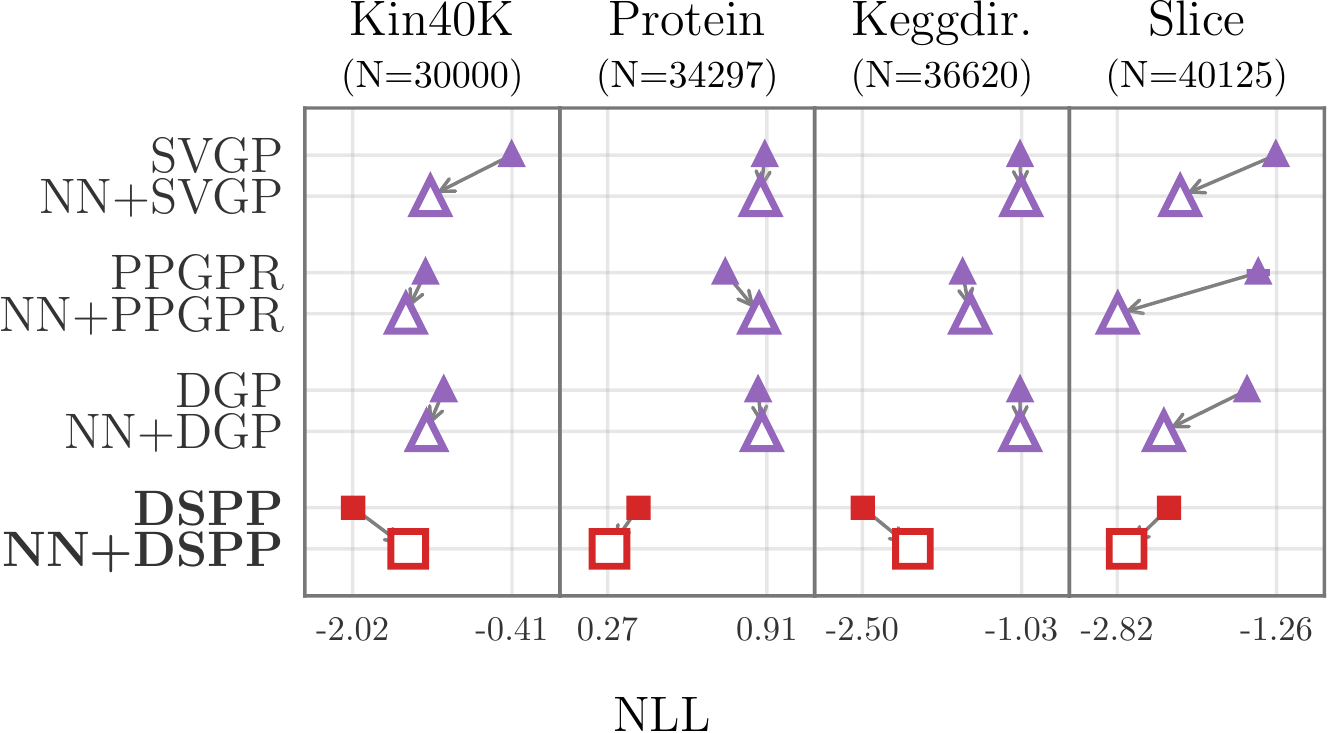}
  \hspace{0.03\linewidth}
  \includegraphics[width=0.48\linewidth]{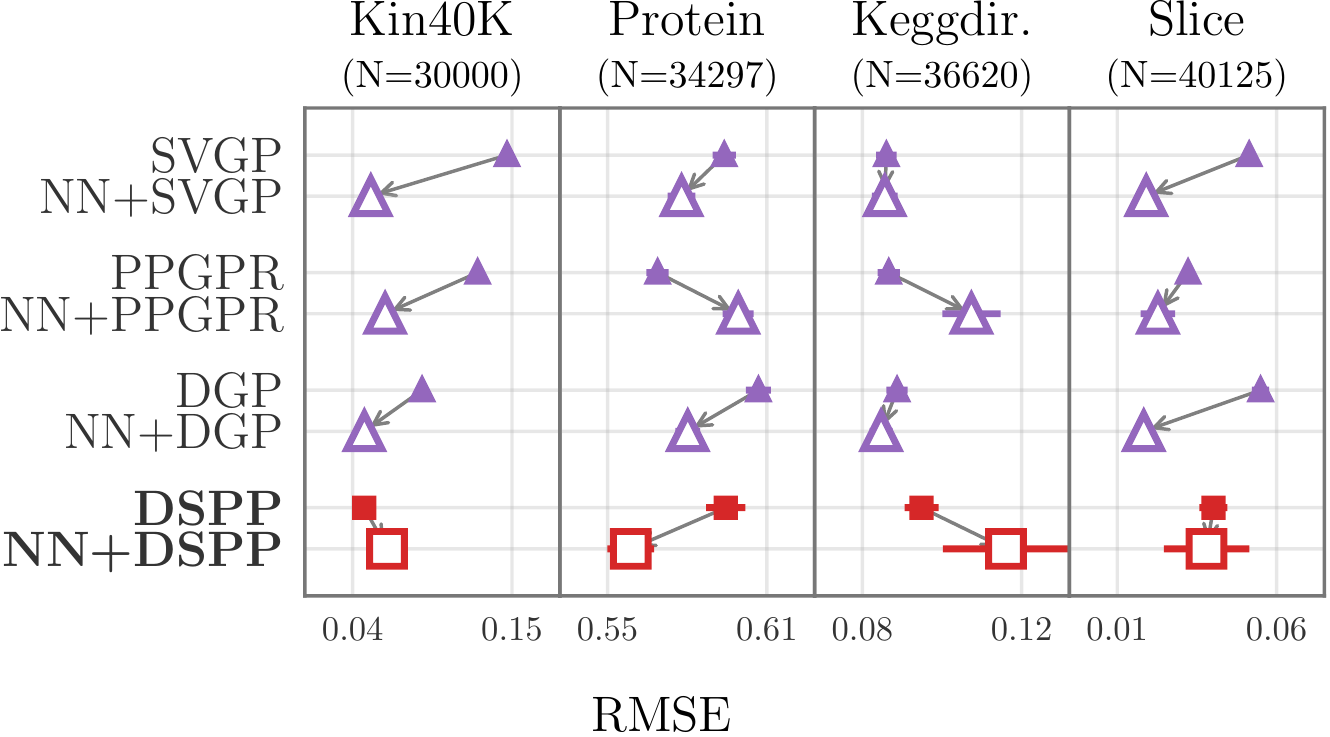}
    \caption{We compare NLLs (left) and RMSEs (right) of neural-network modulated (deep kernel learning) variants of GP/DGP/DSPP models (lower is better).
    Results are presented for 4 of the UCI regression datasets in Sec.~\ref{sec:uci}, averaged over ten random train/test/validation splits. See Sec.~\ref{sec:dkl} for details.}
  \label{fig:dkl}
\end{figure*}

From the previous sets of results we see that DSPP models tend to outperform their single-layer counterparts, both in terms of NLL and RMSE.
A natural question is whether or not these performance improvements can be obtained with other classes of hierarchical models.
More specifically, could we achieve similar results if the hierarchical features are extracted using traditional neural networks rather than GPs?
To answer this question, we consider deep kernel learning (DKL) variants of SVGP, PPRPR, DPG, and DSPP in which the
input features to each model are extracted by a neural network \citep{calandra2016manifold, wilson2016stochastic}.
The neural network parameters are optimized end-to-end alongside the GP/DGP/DSPP parameters.
We conduct experiments on 4 medium-sized UCI regression datasets.
For all DKL models we use a 5-layer neural network architecture proposed by \citet{wilson2016stochastic} to extract features (see Sec.~\ref{sec:suppdkl} for details).

In Fig.~\ref{fig:dkl} we compare the neural network modulated models (denoted by ``NN+'') with their standard counterparts.
We find that adding a neural network to DSPP models has limited impact on performance, improving on some datasets and not
on others.
For SVGP/PPGPR/DGP models, neural networks improve RMSE but have mixed effects on NLL.
On three of the four datasets, none of the NN + SVGP/PPGPR/DGP models matches the NLL of the standard DSPP.
This is particularly notable for the NN+PPGPR model, as it only differs from the standard DSPP model in terms of its ``feature extractor'' layer (neural networks versus GP/quadrature).
These results suggests that hidden GP layers in DSPPs extract features that are complementary to those extracted by neural networks, while enjoying favorable regularization properties.

\section{DISCUSSION}
\label{sec:disc}

We motivated Deep Sigma Point Processes as a finite family of parametric distributions whose structure
mirrors the DGP predictive distribution in Eqn.~\ref{eqn:dgppreddist}. 
It would be interesting to consider model variants that retain a continuous mixture distribution as
in Eqn.~\ref{eqn:dgppreddist} while leveraging the flexibility inherent in the learned quadrature rules
described in Sec.~\ref{sec:quadrule}.
Another open question
is whether DSPPs can be fruitfully applied to other likelihoods, for example those that arise in classification. We leave
the exploration of these directions to future work.

\clearpage
\newpage



\subsubsection*{References}

\bibliography{main}

\clearpage

\appendix
\section{EXPERIMENTAL DETAILS}
\label{sec:suppexp}

We use Mat{\'e}rn kernels with independent length scales for each input dimension throughout.
Throughout we discard input or output dimensions that have negligible variance.

\subsection{QUADRATURE RULE ABLATION STUDY}
\label{sec:suppqr}

The experimental procedure for this experiment follows that described in the next section, with
the difference that the quadrature rule ablation study described in Sec.~\ref{sec:quadruleablation}
only makes use of the 8 smallest UCI datasets and we consider $W \in \{3,4\}$.

\subsection{UNIVARIATE REGRESSION}
\label{sec:suppuci}

We follow the experimental procedure outlined in \citep{jankowiak2019sparse}.
In particular
we use the Adam optimizer for optimization with an initial learning rate of $\ell=0.01$ that is progressively
decreased during the course of training \citep{kingma2014adam}. We use a mini-batch size
of $B=2000$ for the Buzz, Song, 3droad and Houseelectric datasets and $B=10^3$ for all other datasets.
We train for 400 epochs except for the Houseelectric dataset where we train for 150 epochs
and the Buzz, Song, and 3droad datasets where we train for 250 epochs.
We do 10 train/test/validation splits on all datasets, always in the proportion 15:3:2.
All datasets are standardized in both input and output space.
For all 2-layer models we use $M=300$ inducing points for each GP; we initialize with kmeans.
For the DSPP we use quadrature rule QR3 with S=10, while
for the DGP and $\gamma$-DGP we use 10 Monte Carlo samples to approximate the ELBO training objective.
We use the validation set to determine a small set of hyperparameters.
In particular for the DGP we search over $\betareg \in \{0.1, 0.3, 0.5, 1.0\}$ (where, as elsewhere,
$\betareg$ is a constant that scales the KL regularization term). For the $\gamma$-DGP
we search over $\gamma \in \{1.01, 1.03, 1.05, 1.1\}$ (with $\betareg=1$).
For the DSPP we search over $\betareg \in \{0.01, 0.05, 0.2, 1.0\}$.
For all 2-layer models we also search over the hyperparameter $W \in \{3, 5\}$, which controls the
width of the first layer.
For all 2-layer models the mean function in the first layer is linear with learned weights, and the
mean function in the second (final) layer is constant with a learned mean.
As mentioned in the main text, for the DSPP we use diagonal covariance matrices $\bS$ to define variance
functions. In contrast, for the DGP and $\gamma$-DGP we use full rank covariance matrices parameterized
by Cholesky factors, as in the original references.
To evaluate log likelihoods for the DGP and $\gamma$-DGP we use a Monte Carlo estimator with $32$ samples.
For all models we do multiple restarts (3) and only train the best initialization to completion (as judged by training NLL);
this makes the optimization more robust.
As mentioned in the main text, results for PPGPR and OD-SVGP are taken from \citep{jankowiak2019sparse}.

\subsection{MULTIVARIATE REGRESSION}
\label{sec:suppmulti}

We first describe the linear model of coregionalization (LMC) model
structure used by all our multivariate models. We focus on the topmost layer of GPs
of width $W^\prime$, as the deeper layers (here only $\bG$, since this is a 2-layer model) are structured identically
as in the univariate case. Our models
are specified by the generative process
\begin{equation}
\begin{split}
    &\bG \sim p(\cdot | \bX)  \qquad [{\rm GP \,prior \,for \, first \, layer}]  \\
    &\bF \sim p(\cdot | \bG)   \,\qquad [{\rm GP \,prior \,for \, second \, layer}]  \\
    &\bY \sim p(\cdot | \bF \bA , \bSig_{\rm obs}) \qquad [{\rm likelihood}]
\end{split}
\end{equation}
where $\bF$ represents a $N \times W^\prime$-dimensional matrix of GP latent function values,
$\bY$ is a $N \times D_Y$-dimensional matrix of outputs and $\bX$ is the set of $D_X$-dimensional inputs.
Here $\bA$ is a $W^\prime \times D_Y$ matrix of learned coefficients that mixes the topmost GPs.
The likelihood $p(\bY| \cdot)$ is a Normal likelihood with a (block-)diagonal covariance matrix $\bSig_{\rm obs}$
specified by $D_Y$ learnable parameters. Throughout our experiments we choose $W^\prime = D_Y$ and treat $W$ (the width of the first GP layer)
as a hyperparameter.

Our experimental procedure for the multivariate regression experiments follows that of the univariate
regression experiments described in the previous section, with the following differences. For the DSPP
we use $S=8$ quadrature points. For the 2-layer models we use $M=150$ inducing points, while for the single-layer
models we use $M=300$ inducing points. We use a mini-batch size of $B=400$ for all datasets and train for 300 epochs.
To evaluate log likelihoods for the DGP and $\gamma$-DGP we use a Monte Carlo estimator with $16$ samples.
We use 5 random train/test/validation splits for each dataset.

\subsection{MULTILAYER MODELS}
\label{sec:suppmultilayer}

The experimental procedure used for the multilayer experiments in Sec.~\ref{sec:multilayer}
follows the procedure described in Sec.~\ref{sec:suppuci}, with the following differences.
For 3-layer models we use 5 instead of 10 random train/test/validation splits. In addition
to searching over $\betareg$ and the layer width $W$\footnote{Note that here $W$ is the width of the first as well as the
second layer.} for 3-layer models we also search over several discrete topology choices.
In particular the first layer always uses a linear
mean function and the final layer always uses a constant mean function, but the structure of the
second layer---in particular how it depends on the outputs of the first layer---differs:
\begin{enumerate}
    \item the $2^{\rm nd}$ layer mean function is linear and depends on the outputs of the $1^{\rm st}$ layer (i.e.~vanilla feedforward)
    \item the $2^{\rm nd}$ layer mean function is linear and depends on the inputs $\bx$ (but the kernel function only depends on the outputs of the $1^{\rm st}$ layer)
    \item the $2^{\rm nd}$ layer mean function is linear and depends on the outputs of the $1^{\rm st}$ layer \emph{and} the inputs $\bx$ (but the kernel function only depends on the outputs of the $1^{\rm st}$ layer)
    \item the $2^{\rm nd}$ layer mean function is linear and both the mean function and kernel function depend on the inputs $\bx$
        and the outputs of the $1^{\rm st}$ layer
\end{enumerate}
Note that this search over topologies applies to both DSPPs and DGPs.

\subsection{DEEP KERNEL LEARNING REGRESSION}
\label{sec:suppdkl}

The neural network variants of SVGP / PPGPR / DGP / DSPP all use the same 5-layer feature extractor proposed by \citet{wilson2016stochastic}.
The layers have $1000$, $1000$, $500$, $50$, and $20$ hidden units (respectively).
The inputs to the SVGP / PPGPR / DGP / DSPP models are the $20$-dimensional extracted features.
We apply batch normalization \citep{ioffe2015batch} and a ReLU non-linearity after the first four layers.
We z-score the final set of extracted features (out of the $d=20$ layer), which we accomplish using a batch normalization layer without any learned affine transformation.
The neural network parameters are trained jointly with the SVGP / PPGPR / DGP / DSPP parameters using the Adam optimizer.
We apply weight decay only to the neural network parameters.
For all models and datasets, we use the validation set to search over the $\betareg$ regularization parameter
$\betareg \in \{0.01, 0.2, 0.5, 1.0\}$ and the amount of weight decay $\in \{ 10^{-3}, 10^{-4} \}$.
For the DGP/DSPP models we only consider 2-layer models with $W = 5$.
The rest of the training details (learning rate, number of epochs, mini-batch sizes, etc.) match those outlined in Sec.~\ref{sec:suppuci}.

\section{TIME AND SPACE COMPLEXITY}
\label{sec:complexity}

We briefly describe the time and space complexity of 2-layer univariate DSPP models that utilize quadrature rule QR3.
(Extending to deeper and multivariate models is straightforward.)
Our analysis is similar to that of doubly-stochastic Deep Gaussian Processes \citep{salimbeni2017doubly}. In particular, when using QR3, the running time
and space complexities of DSPP are identical to that of the doubly stochastic DGP if the number of quadrature sites $S$ for DSPP is taken to be equal to the number of
samples used for the DGP.

We first note that computing the marginal distribution $q(f(\bx))$ of a \emph{single-layer} sparse Gaussian Process---as in Eqn.~\ref{eqn:qfdist}---is $\mathcal O(M^3)$.
Both the predictive mean (Eqn.~\ref{eqn:meanfunc}) and variance (Eqn.~\ref{eqn:fvar}) require computing $\bK_{MM}^{-1}$ which is a cubic operation.
After this operation all other terms can be computed in $\mathcal O(M^2)$ time.
If the inverse (or, more practically, its Cholesky factor) is cached, all subsequent predictive distributions also require $\mathcal O(M^2)$ time.

\paragraph{Training complexity.}
Each DSPP training iteration requires computing Eqn.~\ref{eqn:dsppobj}.
Again, let $W$ be the width of the hidden GP layer, $S$ be the number of quadrature points, and $M$ be the number of inducing points (which, for simplicity, we assume is the same for both layers).
The right term in Eqn.~\ref{eqn:dsppobj} is the sum of $W + 1$ KL divergences between $M$-dimensional multivariate Normal distributions (one for each hidden-layer GP and one for the final-layer GP), for a total of $\mathcal O(W M^3)$ computation.
To compute the log likelihood term, we compute $q(g_1(\bx)), \ldots, q(g_W(\bx))$, $S$ quadrature points $\hat \bgg_1, \ldots, \hat \bgg_S$, and $q(f(\hat \bgg_1)), \ldots, q(f(\hat \bgg_S))$.
This requires $\mathcal O(W M^3)$ computation for the $\bK_{MM}^{-1}$ terms of each GP, and then multiplies against $W$ matrices of size $M \times B$, where $B$ is the minibatch size.
In total the computational complexity is $\mathcal O( W M^3 + (S+B)WM^2)$.
The space complexity is $\mathcal O( W M^2 + (S+W)MB)$---the size of storing each $\bK_{MM}^{-1}$ as well as the additional vectors in Eqns.~\ref{eqn:meanfunc} and \ref{eqn:fvar} for each marginal distribution $q(f(\hat \bgg_i))$.

\paragraph{Prediction complexity.}
We use nearly the same set of computations at test time as we do during training.
The only difference is that, since the parameters are constant, we do not need to recompute the $\bK_{MM}^{-1}$ matrices at every iteration.
Consequentially, after the one-time $\mathcal O(M^3)$ cost to compute these inverses, the remaining terms can be computed in $\mathcal O((S+B)WM^2)$ time.
The space complexity is still $\mathcal O( W M^2 + (S+W)MB)$.

\section{ADDITIONAL RESULTS}
\label{sec:additional}

\begin{figure*}[t!]
  \centering
  \includegraphics[width=\linewidth,center]{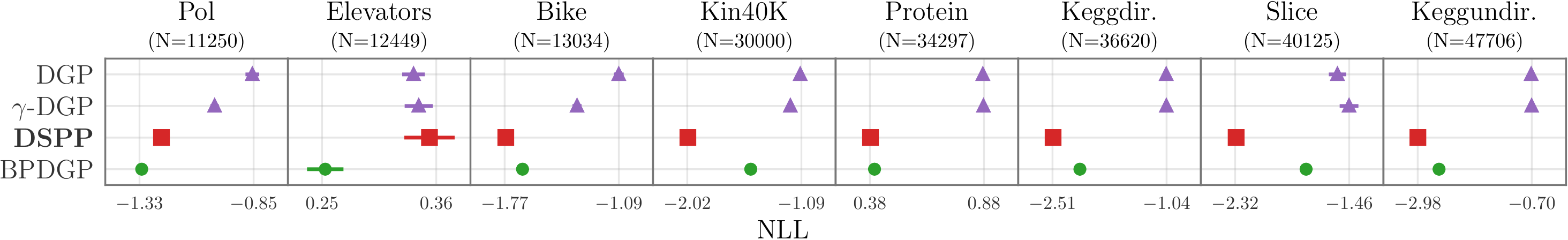}
  \includegraphics[width=\linewidth,center]{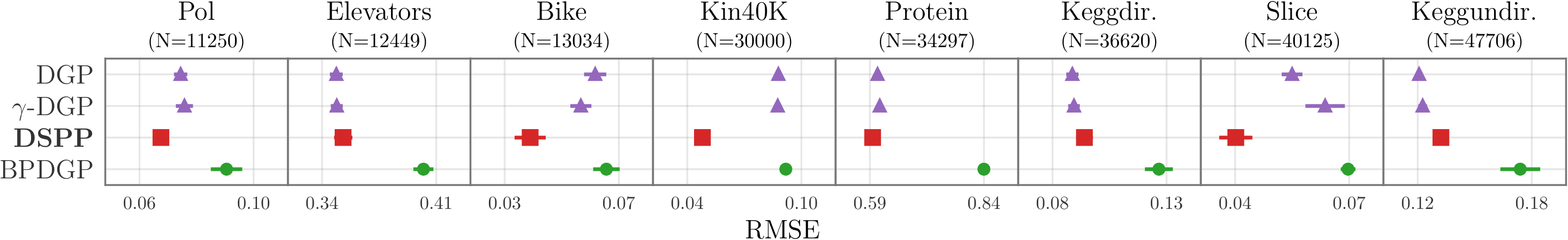}
  \caption{We depict negative log likelihoods (NLL, top) and root mean squared error (RMSE, bottom) for predictive deep GPs trained with biased Monte Carlo sampling (instead of learned quadrature weights).
           Results are averaged over five random train/test/validation splits.}
  \label{fig:biased_results}
\end{figure*}

\begin{figure}[t!]
  \centering
  \includegraphics[width=\linewidth,center]{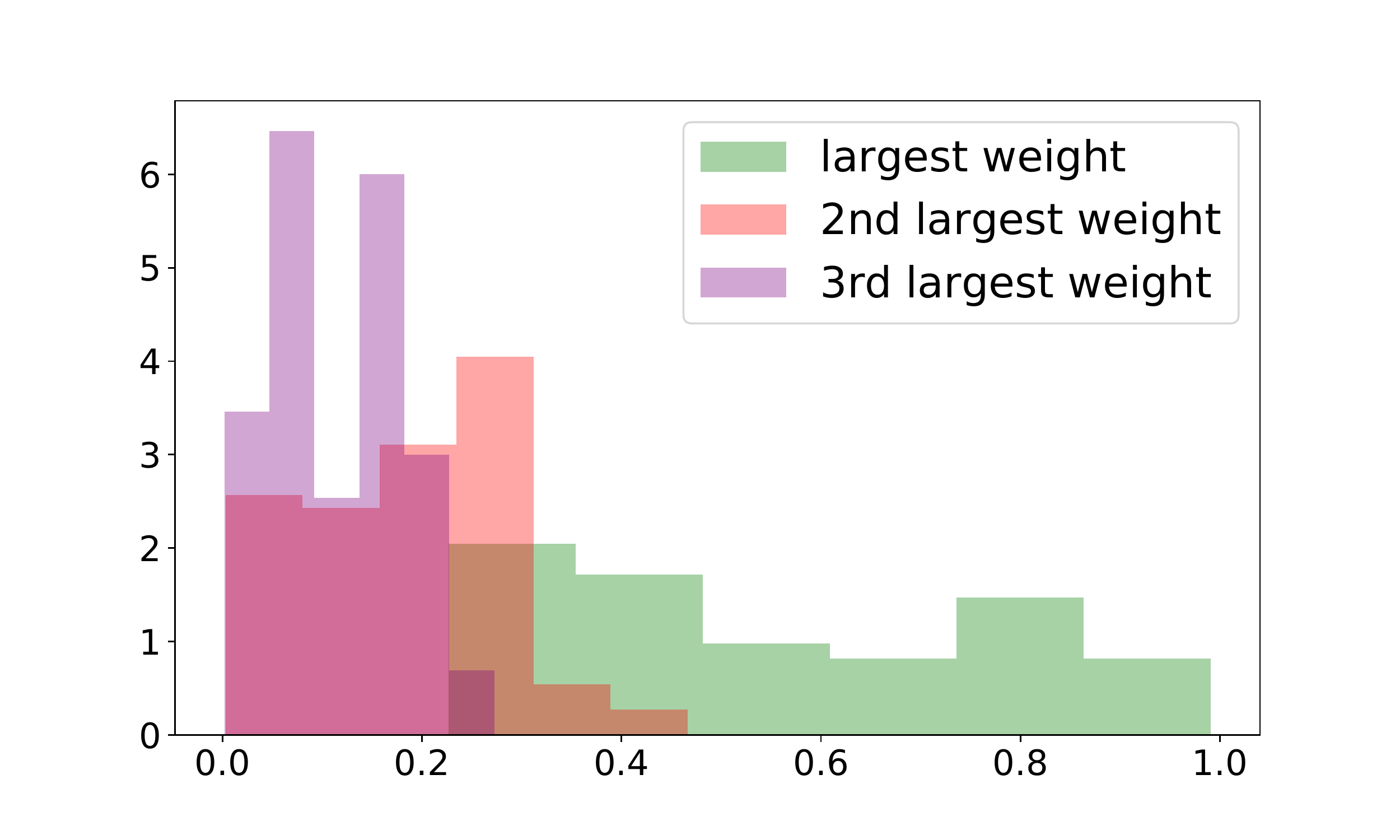}
  \caption{We depict histograms over leading quadrature weights $\omega^{(s)}$ for 96 DSPPs trained on a mixture of univariate regression datasets.}
  \label{fig:omega}
\end{figure}

\paragraph{Quadrature weights.}
In Fig.~\ref{fig:omega} we depict a histogram of learned quadrature weights for 96 DSPPs trained using
quadrature rule QR3.  In particular we follow the experimental procedure described in Sec.~\ref{sec:suppuci}
with $S=10$ quadrature points. We average over 3 train/test/validation splits for the 8 smallest UCI regression datasets with
4 different values of $\betareg$, for a total of 96 model runs. We then depict distributions over the largest, second largest, and third largest
quadrature weights. We see that while a significant amount of the probability mass is put on the leading one or two mixture components,
DSPP training does not result in degenerate weights: the final predictive distribution reflects a diversity of mean and variance functions.

\paragraph{Using MC integration instead of quadrature.}
Rather than using a quadrature scheme or learned weights to evaluate the integral in Eq.~\ref{eqn:dgppreddist}, we could alternatively use Monte Carlo sampling.
As described in Sec.~\ref{sec:dspp}, this would result in a biased estimate of the predictive objective function.
In Fig.~\ref{fig:biased_results} we depict the NLL and RMSE of DSPP models that use biased MC sampling to evaluate Eq.~\ref{eqn:dgppreddist} rather than quadrature.
As this approach bypasses the finite mixture approximation made by DSPP models, we refer to this class of model as Biased Predictive Deep GPs (or {\bf BPDGPs}).
These models are trained following the procedure described in Sec.~\ref{sec:suppuci}.\footnote{
  Integrals are evaluated with 32 MC samples.
}
We find that DSPP models (trained with QR3) tend to outperform the biased BPDGP models, both in terms of NLL and RMSE.
In fact, the biased models achieve even worse RMSE than single-layer SVGP or PPGPR models.

\paragraph{Quadrature ablation study.}
Table~\ref{table:uci_ablation} displays the average NLL, RMSE, and CRPS results across all datasets and train/test/validation splits.

\paragraph{Univariate regression.}
Table~\ref{table:uci} summarizes of Figs.~\ref{fig:ucill}, \ref{fig:ucirmse} and \ref{fig:ucicrps}.
It displays the average NLL, RMSE, and CRPS results across datasets and train/test/validation splits.

\begin{figure}[t!]
  \centering
  \includegraphics[width=\linewidth,center]{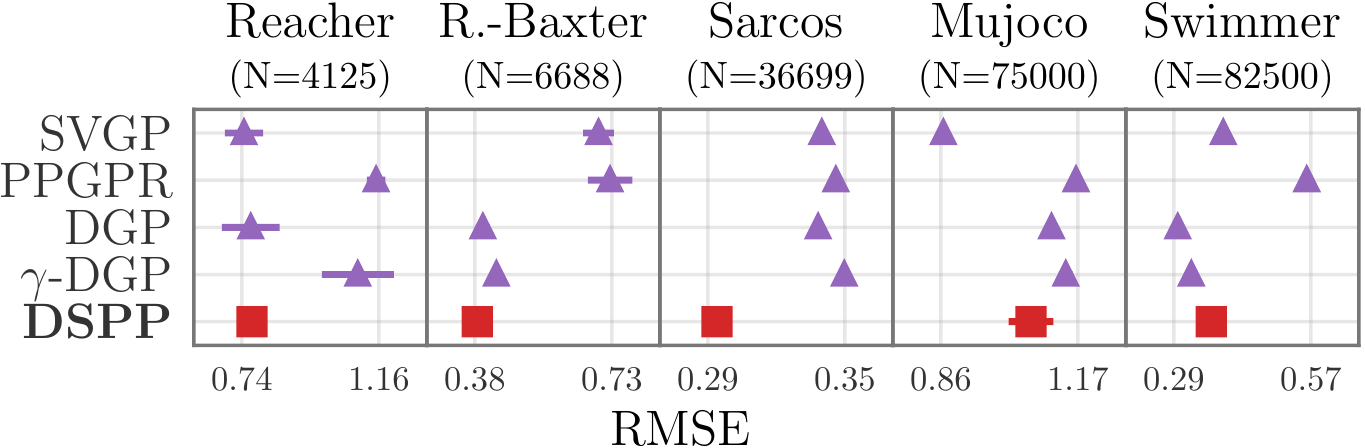}
    \caption{We depict root mean squared errors (RMSEs) for the 5 multivariate regression datasets
             in Sec.~\ref{sec:multi} (lower is better).
    Results are averaged over five random train/test/validation splits.}
  \label{fig:multirmse}
\end{figure}

\paragraph{Multivariate regression.}
Fig.~\ref{fig:multirmse} displays root mean squared error of the multivariate models on all datasets.
Table~\ref{table:multi} summarizes the results of Fig.~\ref{fig:multill_mrmse} and Fig.~\ref{fig:multirmse}.

\begin{figure}[t!]
  \centering
  \includegraphics[width=\linewidth,center]{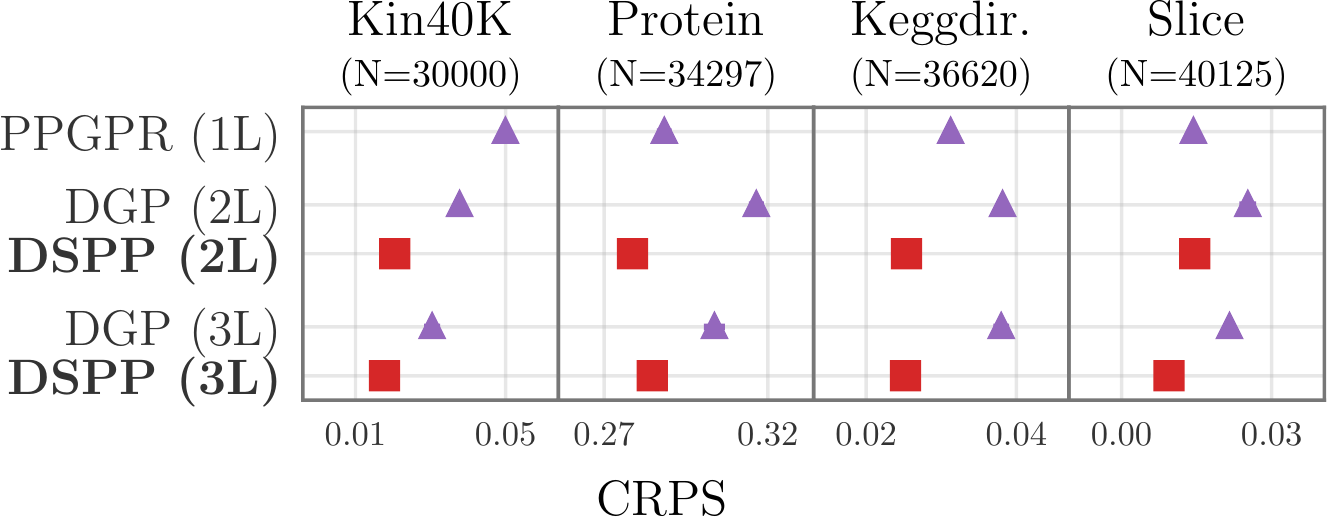}
    \caption{We depict the Continuous Ranked Probability Score (CRPS) for the multi-layer experiment in Sec.~\ref{sec:multilayer} (lower is better).
    Results are averaged over five random train/test/validation splits (for 3-layer models) and ten splits otherwise.}
  \label{fig:threelayercrps}
\end{figure}

\paragraph{Multilayer models.}
Fig.~\ref{fig:threelayercrps} displays the CRPS across 4 datasets (Kin40k, Protein, Keggdirected, and Slice) for models with multiple layers.
Table~\ref{table:uci_multilayer} summarizes the results of Fig.~\ref{fig:threelayerllrmse} and Fig.~\ref{fig:threelayercrps}.

\begin{figure}[t!]
  \centering
  \includegraphics[width=\linewidth,center]{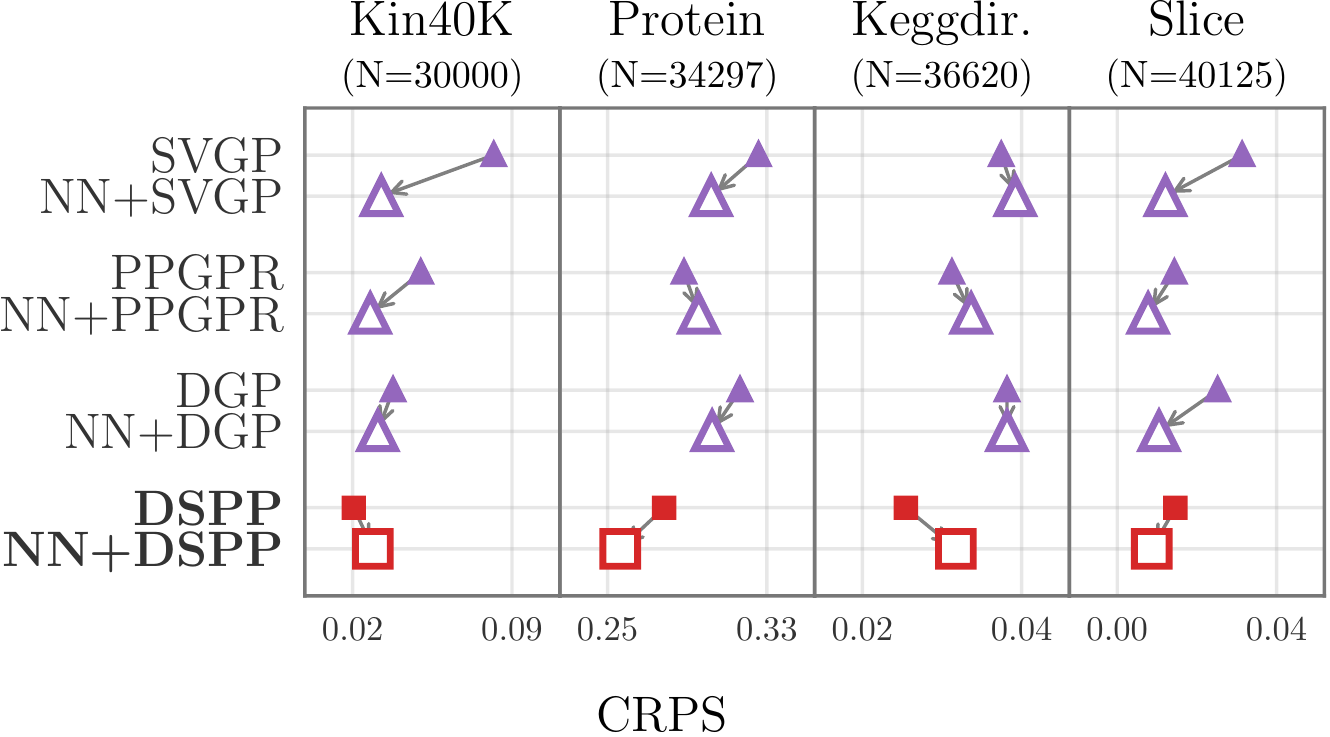}
    \caption{We depict the Continuous Ranked Probability Score (CRPS) of neural-network modulated (deep kernel learning) variants of GP/DGP/DSPP models (lower is better).
    Results are averaged over ten random train/test/validation splits.
		See Sec.~\ref{sec:dkl} for details.}
  \label{fig:dklcrps}
\end{figure}

\paragraph{Comparison with Deep Kernel Learning.}
Fig.~\ref{fig:dklcrps} displays the CRPS across 4 datasets (Kin40k, Protein, Keggdirected, and Slice) for models augmented with neural networks (deep kernel learning).
Table~\ref{table:dkl} summarizes the results of Fig.~\ref{fig:dkl} and Fig.~\ref{fig:dklcrps}.

\paragraph{Results compilations.}
Tables~\ref{table:uci_mega}, \ref{table:multi_mega}, and \ref{table:dkl_mega} report numbers for all experiments in Sec.~\ref{sec:exp}.
Numbers are averages $\pm$ standard errors over train/test/validation splits.

\begin{table}[t!]
  \centering
    \caption{Average NLL, RMSE, and CRPS of different quadrature rules (lower is better).
    Averages are aggregated across the smallest 8 UCI datasets and train/test/validation splits.
    See Sec.~\ref{sec:quadruleablation} for details.
    }
  \resizebox{1.0\linewidth}{!}{%
    \begin{tabular}{ccccccc}
\toprule
{} &   &  DSPP-QR1 &  DSPP-QR2 &             DSPP-QR3 \\
\midrule
NLL  &   &  $-1.460$ &  $-1.444$ &  $\mathbf{ -1.509 }$ \\
RMSE &   &   $0.173$ &   $0.175$ &   $\mathbf{ 0.171 }$ \\
CRPS &   &   $0.077$ &   $0.079$ &   $\mathbf{ 0.076 }$ \\
\bottomrule
\end{tabular}

  }\label{table:uci_ablation}
\end{table}
\begin{table}[t!]
  \centering
    \caption{Average NLL, MRMSE, and RMSE for univariate models (lower is better).
    Averages are aggregated across the 12 univariate UCI datasets and train/test/validation splits.
    See Sec.~\ref{sec:uci} for details.
    }
  \resizebox{1.0\linewidth}{!}{%
    \begin{tabular}{ccccccc}
\toprule
{} &   &   OD-SVGP &     PPGPR &       DGP & $\gamma$-DGP &                 DSPP \\
\midrule
NLL  &   &  $-0.383$ &  $-0.730$ &  $-0.450$ &     $-0.485$ &  $\mathbf{ -1.198 }$ \\
RMSE &   &   $0.242$ &   $0.237$ &   $0.236$ &      $0.238$ &   $\mathbf{ 0.231 }$ \\
CRPS &   &   $0.130$ &   $0.117$ &   $0.124$ &      $0.122$ &   $\mathbf{ 0.111 }$ \\
\bottomrule
\end{tabular}

  }\label{table:uci}
\end{table}
\begin{table}[t!]
  \centering
    \caption{Average NLL, MRMSE, and RMSE for multivariate models (lower is better).
    Averages are aggregated across the five multivariate datasets and train/test/validation splits.
    See Sec.~\ref{sec:multi} for details.
    }
  \resizebox{1.0\linewidth}{!}{%
    \begin{tabular}{ccccccc}
\toprule
{} &   &      SVGP &     PPGPR & $\gamma$-DGP &       DGP &                 DSPP \\
\midrule
NLL   &   &  $-0.179$ &  $-0.889$ &     $-0.379$ &  $-0.240$ &  $\mathbf{ -1.058 }$ \\
MRMSE &   &   $0.198$ &   $0.255$ &      $0.213$ &   $0.188$ &   $\mathbf{ 0.185 }$ \\
RMSE  &   &   $0.608$ &   $0.790$ &      $0.670$ &   $0.583$ &   $\mathbf{ 0.576 }$ \\
\bottomrule
\end{tabular}

  }\label{table:multi}
\end{table}
\begin{table}[t!]
  \centering
    \caption{Average NLL, RMSE, and CRPS for multi-layer models (lower is better).
    Averages are aggregated across 4 of the medium-sized UCI datasets (Kin40K, Protein, Keggdirected, Slice) and train/test/validation splits.
    See Sec.~\ref{sec:multilayer} for details.
    }
  \resizebox{1.0\linewidth}{!}{%
    \begin{tabular}{ccccccc}
\toprule
{} &   & PPGPR (1L) &  DGP (2L) & DSPP (2L) &  DGP (3L) &            DSPP (3L) \\
\midrule
NLL  &   &   $-0.889$ &  $-0.705$ &  $-1.612$ &  $-0.810$ &  $\mathbf{ -1.737 }$ \\
RMSE &   &    $0.203$ &   $0.210$ &   $0.195$ &   $0.214$ &   $\mathbf{ 0.193 }$ \\
CRPS &   &    $0.096$ &   $0.104$ &   $0.085$ &   $0.098$ &   $\mathbf{ 0.084 }$ \\
\bottomrule
\end{tabular}

  }\label{table:uci_multilayer}
\end{table}
\begin{table}[t!]
  \centering
    \caption{Average NLL, RMSE, and CRPS of neural-network modulated (DKL) model variants.
    Averages are aggregated across the 4 UCI datasets (Kin40k, Protein, Keggdirected, and Slice) and train/test/validation splits.
    See Sec.~\ref{sec:dkl} for details.
    }
  \resizebox{1.0\linewidth}{!}{%
    \begin{tabular}{cccccccccc}
\toprule
{} &   &      SVGP &   NN+SVGP &     PPGPR &  NN+PPGPR &       DGP &              NN+DGP &                 DSPP &             NN+DSPP \\
\midrule
NLL  &   &  $-0.456$ &  $-0.897$ &  $-0.889$ &  $-1.231$ &  $-0.705$ &            $-0.948$ &  $\mathbf{ -1.608 }$ &            $-1.485$ \\
RMSE &   &   $0.219$ &   $0.184$ &   $0.203$ &   $0.198$ &   $0.210$ &  $\mathbf{ 0.183 }$ &              $0.194$ &             $0.194$ \\
CRPS &   &   $0.119$ &   $0.096$ &   $0.096$ &   $0.091$ &   $0.104$ &             $0.096$ &              $0.085$ &  $\mathbf{ 0.081 }$ \\
\bottomrule
\end{tabular}

  }\label{table:dkl}
\end{table}


\begin{table*}[t!]
  \centering
    \caption{A compilation of all UCI results from Secs.~\ref{sec:quadruleablation}, \ref{sec:uci}, and \ref{sec:multilayer}.
				For each metric and dataset we bold the result for the best performing method (lower is better for all metrics). $\pm$ indicates standard error.}
  \resizebox{1.0\linewidth}{!}{%
    \begin{tabular}{cccccccccccccc}
\toprule
     &               &   &                       OD-SVGP &                         PPGPR &             $\gamma$-DGP (2L) &                      DGP (2L) &                      DGP (3L) &   &                  DSPP-QR1 (2L) &                  DSPP-QR2 (2L) &                  DSPP-QR3 (2L) &                  DSPP-QR3 (3L) \\
Metric & Dataset &           &                               &                               &                               &                               &                               &           &                                &                                &                                &                                \\
\midrule
\midrule
NLL & Pol &   &            $-0.723 \pm 0.006$ &            $-1.090 \pm 0.009$ &            $-1.014 \pm 0.009$ &            $-0.855 \pm 0.014$ &                           --- &   &  $\mathbf{ -1.236 \pm 0.005 }$ &             $-1.223 \pm 0.006$ &  $\mathbf{ -1.237 \pm 0.008 }$ &                            --- \\
     & Elevators &   &             $0.448 \pm 0.010$ &             $0.368 \pm 0.011$ &  $\mathbf{ 0.343 \pm 0.007 }$ &  $\mathbf{ 0.338 \pm 0.005 }$ &                           --- &   &   $\mathbf{ 0.346 \pm 0.011 }$ &   $\mathbf{ 0.355 \pm 0.013 }$ &   $\mathbf{ 0.354 \pm 0.012 }$ &                            --- \\
     & Bike &   &            $-0.824 \pm 0.009$ &            $-1.426 \pm 0.010$ &            $-1.339 \pm 0.013$ &            $-1.090 \pm 0.015$ &                           --- &   &  $\mathbf{ -1.732 \pm 0.021 }$ &             $-1.717 \pm 0.021$ &  $\mathbf{ -1.763 \pm 0.014 }$ &                            --- \\
     & Kin40K &   &            $-0.830 \pm 0.004$ &            $-1.284 \pm 0.005$ &            $-1.180 \pm 0.006$ &            $-1.100 \pm 0.004$ &            $-1.292 \pm 0.024$ &   &             $-1.850 \pm 0.034$ &             $-1.813 \pm 0.039$ &             $-2.016 \pm 0.012$ &  $\mathbf{ -2.133 \pm 0.011 }$ \\
     & Protein &   &             $0.892 \pm 0.006$ &             $0.743 \pm 0.008$ &             $0.878 \pm 0.005$ &             $0.875 \pm 0.004$ &             $0.814 \pm 0.005$ &   &   $\mathbf{ 0.395 \pm 0.009 }$ &              $0.434 \pm 0.012$ &   $\mathbf{ 0.382 \pm 0.006 }$ &              $0.407 \pm 0.014$ \\
     & Keggdir. &   &            $-1.057 \pm 0.018$ &            $-1.575 \pm 0.015$ &            $-1.043 \pm 0.018$ &            $-1.045 \pm 0.016$ &            $-1.045 \pm 0.031$ &   &             $-2.454 \pm 0.009$ &             $-2.474 \pm 0.011$ &             $-2.503 \pm 0.016$ &  $\mathbf{ -2.556 \pm 0.029 }$ \\
     & Slice &   &            $-1.673 \pm 0.013$ &            $-1.438 \pm 0.057$ &            $-1.461 \pm 0.035$ &            $-1.549 \pm 0.033$ &            $-1.719 \pm 0.009$ &   &             $-2.194 \pm 0.074$ &             $-2.146 \pm 0.069$ &             $-2.312 \pm 0.019$ &  $\mathbf{ -2.666 \pm 0.020 }$ \\
     & Keggundir. &   &            $-0.712 \pm 0.006$ &            $-1.801 \pm 0.013$ &            $-0.703 \pm 0.006$ &            $-0.712 \pm 0.006$ &                           --- &   &  $\mathbf{ -2.959 \pm 0.013 }$ &  $\mathbf{ -2.964 \pm 0.012 }$ &  $\mathbf{ -2.976 \pm 0.020 }$ &                            --- \\
     & 3Droad &   &             $0.231 \pm 0.014$ &            $-0.297 \pm 0.003$ &             $0.216 \pm 0.003$ &             $0.241 \pm 0.003$ &                           --- &   &                            --- &                            --- &  $\mathbf{ -0.488 \pm 0.006 }$ &                            --- \\
     & Song &   &             $1.168 \pm 0.001$ &             $1.103 \pm 0.001$ &             $1.164 \pm 0.001$ &             $1.162 \pm 0.001$ &                           --- &   &                            --- &                            --- &   $\mathbf{ 0.670 \pm 0.045 }$ &                            --- \\
     & Buzz &   &             $0.044 \pm 0.002$ &            $-0.047 \pm 0.001$ &             $0.002 \pm 0.002$ &             $0.001 \pm 0.002$ &                           --- &   &                            --- &                            --- &  $\mathbf{ -0.209 \pm 0.014 }$ &                            --- \\
     & Houseelectric &   &            $-1.559 \pm 0.002$ &            $-2.020 \pm 0.003$ &            $-1.686 \pm 0.003$ &            $-1.671 \pm 0.003$ &                           --- &   &                            --- &                            --- &  $\mathbf{ -2.281 \pm 0.003 }$ &                            --- \\
\midrule
RMSE & Pol &   &             $0.109 \pm 0.001$ &             $0.077 \pm 0.001$ &             $0.076 \pm 0.002$ &             $0.074 \pm 0.001$ &                           --- &   &   $\mathbf{ 0.065 \pm 0.001 }$ &   $\mathbf{ 0.064 \pm 0.002 }$ &              $0.067 \pm 0.001$ &                            --- \\
     & Elevators &   &             $0.370 \pm 0.003$ &             $0.361 \pm 0.003$ &  $\mathbf{ 0.349 \pm 0.002 }$ &  $\mathbf{ 0.349 \pm 0.002 }$ &                           --- &   &   $\mathbf{ 0.350 \pm 0.002 }$ &   $\mathbf{ 0.351 \pm 0.003 }$ &   $\mathbf{ 0.353 \pm 0.003 }$ &                            --- \\
     & Bike &   &             $0.097 \pm 0.002$ &             $0.060 \pm 0.001$ &             $0.057 \pm 0.002$ &             $0.062 \pm 0.002$ &                           --- &   &   $\mathbf{ 0.043 \pm 0.004 }$ &   $\mathbf{ 0.044 \pm 0.004 }$ &   $\mathbf{ 0.039 \pm 0.003 }$ &                            --- \\
     & Kin40K &   &             $0.109 \pm 0.001$ &             $0.126 \pm 0.001$ &             $0.088 \pm 0.001$ &             $0.088 \pm 0.000$ &             $0.075 \pm 0.003$ &   &              $0.057 \pm 0.002$ &              $0.060 \pm 0.003$ &              $0.048 \pm 0.001$ &   $\mathbf{ 0.042 \pm 0.001 }$ \\
     & Protein &   &             $0.591 \pm 0.003$ &  $\mathbf{ 0.569 \pm 0.002 }$ &             $0.612 \pm 0.002$ &             $0.607 \pm 0.002$ &             $0.643 \pm 0.005$ &   &              $0.599 \pm 0.003$ &              $0.609 \pm 0.002$ &              $0.596 \pm 0.003$ &              $0.606 \pm 0.003$ \\
     & Keggdir. &   &  $\mathbf{ 0.085 \pm 0.001 }$ &  $\mathbf{ 0.087 \pm 0.001 }$ &             $0.089 \pm 0.001$ &             $0.089 \pm 0.001$ &  $\mathbf{ 0.089 \pm 0.003 }$ &   &              $0.092 \pm 0.002$ &              $0.094 \pm 0.002$ &              $0.094 \pm 0.002$ &              $0.094 \pm 0.005$ \\
     & Slice &   &             $0.043 \pm 0.001$ &  $\mathbf{ 0.032 \pm 0.001 }$ &             $0.064 \pm 0.003$ &             $0.055 \pm 0.001$ &             $0.048 \pm 0.001$ &   &              $0.049 \pm 0.007$ &              $0.047 \pm 0.008$ &              $0.040 \pm 0.002$ &   $\mathbf{ 0.030 \pm 0.004 }$ \\
     & Keggundir. &   &  $\mathbf{ 0.119 \pm 0.001 }$ &             $0.123 \pm 0.001$ &             $0.123 \pm 0.001$ &             $0.121 \pm 0.001$ &                           --- &   &              $0.132 \pm 0.001$ &              $0.132 \pm 0.001$ &              $0.132 \pm 0.001$ &                            --- \\
     & 3Droad &   &             $0.303 \pm 0.004$ &             $0.304 \pm 0.001$ &             $0.322 \pm 0.001$ &             $0.322 \pm 0.001$ &                           --- &   &                            --- &                            --- &   $\mathbf{ 0.296 \pm 0.002 }$ &                            --- \\
     & Song &   &             $0.778 \pm 0.001$ &  $\mathbf{ 0.770 \pm 0.001 }$ &             $0.782 \pm 0.001$ &             $0.780 \pm 0.001$ &                           --- &   &                            --- &                            --- &              $0.820 \pm 0.008$ &                            --- \\
     & Buzz &   &             $0.256 \pm 0.001$ &             $0.283 \pm 0.001$ &  $\mathbf{ 0.244 \pm 0.001 }$ &  $\mathbf{ 0.244 \pm 0.000 }$ &                           --- &   &                            --- &                            --- &              $0.247 \pm 0.001$ &                            --- \\
     & Houseelectric &   &             $0.050 \pm 0.000$ &             $0.046 \pm 0.000$ &             $0.046 \pm 0.000$ &             $0.046 \pm 0.000$ &                           --- &   &                            --- &                            --- &   $\mathbf{ 0.042 \pm 0.000 }$ &                            --- \\
\midrule
CRPS & Pol &   &             $0.059 \pm 0.000$ &             $0.040 \pm 0.000$ &             $0.040 \pm 0.000$ &             $0.045 \pm 0.001$ &                           --- &   &   $\mathbf{ 0.033 \pm 0.000 }$ &              $0.034 \pm 0.000$ &   $\mathbf{ 0.034 \pm 0.000 }$ &                            --- \\
     & Elevators &   &             $0.203 \pm 0.001$ &             $0.195 \pm 0.002$ &  $\mathbf{ 0.186 \pm 0.001 }$ &  $\mathbf{ 0.186 \pm 0.001 }$ &                           --- &   &              $0.189 \pm 0.001$ &              $0.190 \pm 0.001$ &              $0.191 \pm 0.001$ &                            --- \\
     & Bike &   &             $0.051 \pm 0.001$ &             $0.028 \pm 0.000$ &             $0.027 \pm 0.000$ &             $0.034 \pm 0.001$ &                           --- &   &   $\mathbf{ 0.021 \pm 0.001 }$ &              $0.021 \pm 0.001$ &   $\mathbf{ 0.019 \pm 0.001 }$ &                            --- \\
     & Kin40K &   &             $0.056 \pm 0.000$ &             $0.050 \pm 0.000$ &             $0.036 \pm 0.000$ &             $0.038 \pm 0.000$ &             $0.030 \pm 0.001$ &   &              $0.024 \pm 0.001$ &              $0.025 \pm 0.001$ &              $0.020 \pm 0.000$ &   $\mathbf{ 0.018 \pm 0.000 }$ \\
     & Protein &   &             $0.317 \pm 0.001$ &             $0.288 \pm 0.001$ &             $0.315 \pm 0.001$ &             $0.317 \pm 0.001$ &             $0.304 \pm 0.002$ &   &   $\mathbf{ 0.281 \pm 0.002 }$ &              $0.288 \pm 0.001$ &   $\mathbf{ 0.279 \pm 0.002 }$ &              $0.285 \pm 0.002$ \\
     & Keggdir. &   &             $0.037 \pm 0.000$ &             $0.031 \pm 0.000$ &             $0.037 \pm 0.000$ &             $0.038 \pm 0.000$ &             $0.038 \pm 0.001$ &   &   $\mathbf{ 0.025 \pm 0.000 }$ &   $\mathbf{ 0.026 \pm 0.000 }$ &   $\mathbf{ 0.025 \pm 0.000 }$ &   $\mathbf{ 0.025 \pm 0.001 }$ \\
     & Slice &   &             $0.021 \pm 0.000$ &             $0.014 \pm 0.000$ &             $0.026 \pm 0.000$ &             $0.025 \pm 0.001$ &             $0.022 \pm 0.000$ &   &              $0.019 \pm 0.003$ &              $0.019 \pm 0.003$ &              $0.015 \pm 0.000$ &   $\mathbf{ 0.009 \pm 0.000 }$ \\
     & Keggundir. &   &             $0.051 \pm 0.000$ &             $0.036 \pm 0.000$ &             $0.050 \pm 0.000$ &             $0.051 \pm 0.000$ &                           --- &   &   $\mathbf{ 0.027 \pm 0.000 }$ &   $\mathbf{ 0.027 \pm 0.000 }$ &   $\mathbf{ 0.027 \pm 0.000 }$ &                            --- \\
     & 3Droad &   &             $0.163 \pm 0.002$ &             $0.138 \pm 0.000$ &             $0.161 \pm 0.000$ &             $0.164 \pm 0.001$ &                           --- &   &                            --- &                            --- &   $\mathbf{ 0.128 \pm 0.001 }$ &                            --- \\
     & Song &   &             $0.434 \pm 0.000$ &  $\mathbf{ 0.422 \pm 0.000 }$ &             $0.433 \pm 0.000$ &             $0.432 \pm 0.000$ &                           --- &   &                            --- &                            --- &              $0.448 \pm 0.005$ &                            --- \\
     & Buzz &   &             $0.135 \pm 0.000$ &             $0.134 \pm 0.000$ &             $0.129 \pm 0.000$ &             $0.130 \pm 0.000$ &                           --- &   &                            --- &                            --- &   $\mathbf{ 0.126 \pm 0.000 }$ &                            --- \\
     & Houseelectric &   &             $0.026 \pm 0.000$ &             $0.022 \pm 0.000$ &             $0.023 \pm 0.000$ &             $0.023 \pm 0.000$ &                           --- &   &                            --- &                            --- &   $\mathbf{ 0.018 \pm 0.000 }$ &                            --- \\
\bottomrule
\end{tabular}

  }\label{table:uci_mega}
\end{table*}
\begin{table*}[t!]
  \centering
    \caption{A compilation of all multivariate results from Sec.~\ref{sec:multi}.
				For each metric and dataset we bold the result for the best performing method (lower is better for all metrics). $\pm$ indicates standard error.}
  \resizebox{0.62\linewidth}{!}{%
    \begin{tabular}{cccccccccc}
\toprule
     &         &   &                          SVGP &               PPGPR &   $\gamma$-DGP (2L) &                      DGP (2L) &   &                  DSPP-QR3 (2L) \\
Metric & Dataset &           &                               &                     &                     &                               &           &                                \\
\midrule
\midrule
NLL & Reacher &   &             $0.460 \pm 0.048$ &  $-0.454 \pm 0.005$ &   $0.215 \pm 0.032$ &             $0.484 \pm 0.032$ &   &  $\mathbf{ -0.491 \pm 0.004 }$ \\
     & R.-Baxter &   &             $0.112 \pm 0.042$ &  $-0.498 \pm 0.010$ &  $-0.728 \pm 0.012$ &            $-0.303 \pm 0.036$ &   &  $\mathbf{ -0.910 \pm 0.005 }$ \\
     & Sarcos &   &            $-0.703 \pm 0.001$ &  $-1.032 \pm 0.002$ &  $-0.700 \pm 0.002$ &            $-0.694 \pm 0.003$ &   &  $\mathbf{ -1.169 \pm 0.003 }$ \\
     & Mujoco &   &             $0.141 \pm 0.002$ &  $-0.563 \pm 0.001$ &   $0.335 \pm 0.007$ &             $0.339 \pm 0.005$ &   &  $\mathbf{ -0.636 \pm 0.010 }$ \\
     & Swimmer &   &            $-0.905 \pm 0.011$ &  $-1.898 \pm 0.002$ &  $-1.018 \pm 0.003$ &            $-1.025 \pm 0.007$ &   &  $\mathbf{ -2.085 \pm 0.002 }$ \\
\midrule
MRMSE & Reacher &   &  $\mathbf{ 0.217 \pm 0.009 }$ &   $0.330 \pm 0.005$ &   $0.323 \pm 0.017$ &  $\mathbf{ 0.237 \pm 0.015 }$ &   &   $\mathbf{ 0.230 \pm 0.007 }$ \\
     & R.-Baxter &   &             $0.247 \pm 0.007$ &   $0.263 \pm 0.010$ &   $0.140 \pm 0.001$ &  $\mathbf{ 0.126 \pm 0.003 }$ &   &   $\mathbf{ 0.121 \pm 0.003 }$ \\
     & Sarcos &   &             $0.124 \pm 0.000$ &   $0.126 \pm 0.000$ &   $0.128 \pm 0.000$ &             $0.123 \pm 0.000$ &   &   $\mathbf{ 0.108 \pm 0.001 }$ \\
     & Mujoco &   &  $\mathbf{ 0.286 \pm 0.001 }$ &   $0.383 \pm 0.001$ &   $0.373 \pm 0.002$ &             $0.362 \pm 0.002$ &   &              $0.352 \pm 0.008$ \\
     & Swimmer &   &             $0.115 \pm 0.001$ &   $0.172 \pm 0.000$ &   $0.100 \pm 0.001$ &  $\mathbf{ 0.092 \pm 0.001 }$ &   &              $0.114 \pm 0.004$ \\
\midrule
RMSE & Reacher &   &  $\mathbf{ 0.747 \pm 0.029 }$ &   $1.151 \pm 0.014$ &   $1.095 \pm 0.055$ &  $\mathbf{ 0.767 \pm 0.044 }$ &   &   $\mathbf{ 0.771 \pm 0.015 }$ \\
     & R.-Baxter &   &             $0.696 \pm 0.020$ &   $0.725 \pm 0.028$ &   $0.435 \pm 0.005$ &  $\mathbf{ 0.401 \pm 0.009 }$ &   &   $\mathbf{ 0.386 \pm 0.010 }$ \\
     & Sarcos &   &             $0.340 \pm 0.001$ &   $0.346 \pm 0.001$ &   $0.350 \pm 0.001$ &             $0.338 \pm 0.001$ &   &   $\mathbf{ 0.294 \pm 0.001 }$ \\
     & Mujoco &   &  $\mathbf{ 0.866 \pm 0.002 }$ &   $1.166 \pm 0.004$ &   $1.142 \pm 0.008$ &             $1.110 \pm 0.006$ &   &              $1.064 \pm 0.025$ \\
     & Swimmer &   &             $0.391 \pm 0.002$ &   $0.561 \pm 0.001$ &   $0.326 \pm 0.001$ &  $\mathbf{ 0.298 \pm 0.003 }$ &   &              $0.366 \pm 0.012$ \\
\bottomrule
\end{tabular}

  }\label{table:multi_mega}
\end{table*}
\begin{table*}[t!]
  \centering
    \caption{A compilation of all deep kernel learning results (UCI datasets) from Sec.~\ref{sec:dkl}.
				For each metric and dataset we bold the result for the best performing method (lower is better for all metrics). $\pm$ indicates standard error.}
  \resizebox{0.87\linewidth}{!}{%
    \begin{tabular}{ccccccccccccc}
\toprule
     &       &   &                          SVGP &                       NN+SVGP &                         PPGPR &                       NN+PPGPR &            DGP (2L) &                        NN+DGP &   &                           DSPP &                       NN+DSPP \\
Metric & Dataset &           &                               &                               &                               &                                &                     &                               &           &                                &                               \\
\midrule
\midrule
NLL & Kin40K &   &            $-0.414 \pm 0.002$ &            $-1.235 \pm 0.004$ &            $-1.284 \pm 0.005$ &             $-1.483 \pm 0.002$ &  $-1.100 \pm 0.004$ &            $-1.273 \pm 0.005$ &   &  $\mathbf{ -2.016 \pm 0.012 }$ &            $-1.458 \pm 0.011$ \\
     & Protein &   &             $0.902 \pm 0.003$ &             $0.885 \pm 0.005$ &             $0.743 \pm 0.008$ &              $0.879 \pm 0.017$ &   $0.875 \pm 0.004$ &             $0.890 \pm 0.004$ &   &              $0.394 \pm 0.008$ &  $\mathbf{ 0.278 \pm 0.010 }$ \\
     & Keggdir. &   &            $-1.045 \pm 0.017$ &            $-1.035 \pm 0.012$ &            $-1.575 \pm 0.015$ &             $-1.505 \pm 0.015$ &  $-1.045 \pm 0.016$ &            $-1.044 \pm 0.014$ &   &  $\mathbf{ -2.498 \pm 0.015 }$ &            $-2.034 \pm 0.022$ \\
     & Slice &   &            $-1.267 \pm 0.003$ &            $-2.204 \pm 0.022$ &            $-1.438 \pm 0.057$ &  $\mathbf{ -2.815 \pm 0.009 }$ &  $-1.549 \pm 0.033$ &            $-2.363 \pm 0.016$ &   &             $-2.312 \pm 0.019$ &            $-2.728 \pm 0.010$ \\
\midrule
RMSE & Kin40K &   &             $0.147 \pm 0.001$ &             $0.052 \pm 0.001$ &             $0.126 \pm 0.001$ &              $0.062 \pm 0.002$ &   $0.088 \pm 0.000$ &  $\mathbf{ 0.048 \pm 0.001 }$ &   &   $\mathbf{ 0.048 \pm 0.001 }$ &             $0.064 \pm 0.004$ \\
     & Protein &   &             $0.594 \pm 0.002$ &             $0.578 \pm 0.003$ &             $0.569 \pm 0.002$ &              $0.599 \pm 0.003$ &   $0.607 \pm 0.002$ &             $0.580 \pm 0.002$ &   &              $0.595 \pm 0.004$ &  $\mathbf{ 0.559 \pm 0.004 }$ \\
     & Keggdir. &   &  $\mathbf{ 0.086 \pm 0.001 }$ &  $\mathbf{ 0.086 \pm 0.002 }$ &  $\mathbf{ 0.087 \pm 0.001 }$ &              $0.107 \pm 0.004$ &   $0.089 \pm 0.001$ &  $\mathbf{ 0.085 \pm 0.001 }$ &   &              $0.095 \pm 0.002$ &             $0.116 \pm 0.008$ \\
     & Slice &   &             $0.051 \pm 0.001$ &  $\mathbf{ 0.019 \pm 0.001 }$ &             $0.032 \pm 0.001$ &              $0.023 \pm 0.003$ &   $0.055 \pm 0.001$ &  $\mathbf{ 0.018 \pm 0.000 }$ &   &              $0.040 \pm 0.002$ &             $0.038 \pm 0.007$ \\
\midrule
CRPS & Kin40K &   &             $0.082 \pm 0.000$ &             $0.033 \pm 0.000$ &             $0.050 \pm 0.000$ &              $0.028 \pm 0.000$ &   $0.038 \pm 0.000$ &             $0.031 \pm 0.000$ &   &   $\mathbf{ 0.020 \pm 0.000 }$ &             $0.029 \pm 0.001$ \\
     & Protein &   &             $0.326 \pm 0.001$ &             $0.302 \pm 0.002$ &             $0.288 \pm 0.001$ &              $0.296 \pm 0.001$ &   $0.317 \pm 0.001$ &             $0.303 \pm 0.002$ &   &              $0.278 \pm 0.002$ &  $\mathbf{ 0.256 \pm 0.003 }$ \\
     & Keggdir. &   &             $0.037 \pm 0.000$ &             $0.039 \pm 0.000$ &             $0.031 \pm 0.000$ &              $0.034 \pm 0.000$ &   $0.038 \pm 0.000$ &             $0.038 \pm 0.000$ &   &   $\mathbf{ 0.025 \pm 0.000 }$ &             $0.032 \pm 0.001$ \\
     & Slice &   &             $0.031 \pm 0.000$ &             $0.012 \pm 0.000$ &             $0.014 \pm 0.000$ &   $\mathbf{ 0.008 \pm 0.000 }$ &   $0.025 \pm 0.001$ &             $0.010 \pm 0.000$ &   &              $0.015 \pm 0.000$ &             $0.009 \pm 0.000$ \\
\bottomrule
\end{tabular}

  }\label{table:dkl_mega}
\end{table*}

\end{document}